\documentclass[oneside,11pt]{article}

\usepackage{tikz}
\usetikzlibrary{3d,calc}
\usepackage{mathrsfs,placeins,soulutf8}
\usepackage[linesnumbered,ruled]{algorithm2e} 
\usepackage{tikz}

\usepackage[a4paper, total={6in, 8in}]{geometry}

\usepackage{mathtools}
\usetikzlibrary{shapes.geometric, arrows}
\usepackage{algorithmic}
\usepackage{bm}
\usepackage{booktabs}
\usepackage[english]{babel}
\usepackage[utf8]{inputenc}
\usepackage{multirow}
\usepackage{adjustbox}
\usepackage{subcaption,afterpage} 
\usepackage{mathtools}
\usepackage{amsfonts}
\usepackage[colorinlistoftodos]{todonotes}
\usepackage{enumerate}
\usepackage{hyperref}
\RequirePackage{epsfig}
\RequirePackage{amssymb}
\usepackage{amsthm}
\RequirePackage{natbib}
\RequirePackage{graphicx}

\if@usehyper
\RequirePackage[colorlinks=false,allbordercolors={1 1 1}]{hyperref}
\fi

\if@abbrvbib
\else
\fi

\bibpunct{(}{)}{;}{a}{,}{,}
 
\newcommand{\excise}[1]{}

\usepackage{amsmath}
\usepackage{cleveref}
\usepackage{dsfont}

\newcommand\RR{\mathbb{R}}

\newcommand\EE{\mathbb{E}}

\newcommand\MM{\mathcal{M}}

\newcommand\vol{\mathrm{Vol}}
\theoremstyle{plain}
\newtheorem{theorem}{Theorem}[section]

\newtheorem{lemma}[theorem]{Lemma}
\newtheorem{corollary}[theorem]{Corollary}
\theoremstyle{definition}
\newtheorem{definition}[theorem]{Definition}
\newtheorem{assumption}[theorem]{Assumption}
\theoremstyle{remark}

\makeatletter
\newtheorem{rep@theorem}{\rep@title}
\newcommand{\newreptheorem}[2]{%
\newenvironment{rep#1}[1]{%
 \def\rep@title{#2 \ref{##1}}%
 \begin{rep@theorem}}%
 {\end{rep@theorem}}}
\makeatother

\newreptheorem{theorem}{Theorem}
\newreptheorem{lemma}{Lemma}
\usepackage{changepage}
\newif\ifhighlightchanges
\newcommand{\changedreviewerone}[1]{\ifhighlightchanges\textcolor{red}{#1}\else#1\fi}

\newcommand{\changedreviewertwo}[1]{\ifhighlightchanges\textcolor{blue}{#1}\else#1\fi}

\usepackage{lastpage}

\begin{document}
%\firstpageno{1}
\title{Deep Generative Models: Complexity, Dimensionality, and Approximation}

% \author{\name Kevin Wang \email kcwang@unc.edu\\
% 		\addr   Department of Biostatistics\\
% 		University of North Carolina at Chapel Hill\\
%   \AND
%        \name Hongqian Niu \email hong.niu@unc.edu \\
%        \addr   Department of Biostatistics\\
% 		University of North Carolina at Chapel Hill \\
%   \AND
%      \name  Yixin Wang \email yixinw@umich.edu\\
%        \addr Department of Statistics\\
%        University of Michigan \\
%   \AND
%      \name  Didong Li \email didongli@unc.edu \\
%        \addr Department of Biostatistics\\
%        University of North Carolina at Chapel Hill	
% }
\author{
  Kevin Wang$^1$,  Hongqian Niu$^1$, Yixin Wang$^2$, Didong Li$^1$\footnotemark[1]\thanks{didongli@unc.edu}\\
  Department of Biostatistics, University of North Carolina at Chapel Hill$^1$\\ 
  Department of Statistics, University of Michigan$^2$
}
\date{}

\maketitle

\begin{abstract}%   <- trailing '%' for backward compatibility of .sty file
    Generative networks have shown remarkable success in learning complex data distributions, particularly in generating high-dimensional data from lower-dimensional inputs. While this capability is well-documented empirically, its theoretical underpinning remains unclear. One common theoretical explanation appeals to the widely accepted manifold hypothesis, which suggests that many real-world datasets, such as images and signals, often possess intrinsic low-dimensional geometric structures. Under this manifold hypothesis, it is widely believed that to approximate a distribution on a $d$-dimensional Riemannian manifold, the latent dimension needs to be at least $d$ or $d+1$. In this work, we show that this requirement on the latent dimension is not necessary by demonstrating that generative networks can approximate distributions on $d$-dimensional Riemannian manifolds from inputs of any arbitrary dimension, even lower than $d$, taking inspiration from the concept of space-filling curves. This approach, in turn, leads to a super-exponential complexity bound of the deep neural networks through expanded neurons. Our findings thus challenge the conventional belief on the relationship between input dimensionality and the ability of generative networks to model data distributions. This novel insight not only corroborates the practical effectiveness of generative networks in handling complex data structures, but also underscores a critical trade-off between approximation error, dimensionality, and model complexity. 
\end{abstract}


\section{Introduction}

    Generative models, such as generative adversarial networks (GANs, \citealp{goodfellow2014generative}) and variational auto-encoders (VAEs, \citealp{kingma2013autoVAE}), have become a central topic in machine learning. These models are versatile, addressing various problems ranging from image generation~\citep{bao2017cvae,han2018gan}, style transfer~\citep{karras2019styleGAN}, anomaly detection~\citep{xia2022gan}, to data augmentation~\citep{tran2021data}, achieving increasingly accurate results across disciplines. Despite these advances, GANs, in particular, face challenges such as mode collapse, vanishing gradients, and training instability, particularly when distributions are not continuous or have disjoint supports~\citep{arjovsky2017towards}.
    
    While traditional GANs and VAEs primarily relied on the Kullback–Leibler divergence (KL) to measure distances between input and output distributions, more recent approaches have greatly improved training performance. The shift from KL to Wasserstein distance~\citep{villani2009optimal} is a notable example, as exemplified in the Wasserstein GAN~(WGAN, \citealp{arjovsky2017wasserstein}), and further extended to Wasserstein VAE (WVAE, \citealp{ambrogioni2018wasserstein}) and Wasserstein Auto Encoders \citep{tolstikhin2017wasserstein}. Beyond this, there have been substantial enhancements in model architecture, loss functions, and regularization techniques. Advances such as conditional GANs (cGAN, \citealp{mirza2014conditional}) for targeted image generation, auxiliary classifier GANs (AC-GANs, \citealp{odena2017conditional}), and self-attention GANs (SA-GANs, \citealp{zhang2019self}) have enriched the versatility and effectiveness of GANs. Similarly, VAEs have seen improvements with techniques such as hierarchical latent variables~\citep{vahdat2020nvae,sonderby2016ladder} and incorporation of normalizing flows~\citep{kingma2016improved}, refining their ability to model complex distributions. These developments collectively contribute to the generation of high-quality, high-dimensional data from simpler, lower-dimensional distributions, with notable impact in fields such as photorealistic image generation~\citep{wang2018high,sarkar2021humangan}. 

    The theoretical advancements in deep generative models, paralleling their practical applications, have been substantial. In particular, the depth and width of network architectures in these models have been theoretically shown to significantly impact their approximation capabilities, as detailed in the works focused on neural network expressiveness~\citep{eldan2016power}. Furthermore, the challenges of mode collapse in GANs and the propensity of VAEs to produce overly smooth outputs have catalyzed research into improving their architectural designs and training methodologies~\citep{goodfellow2014generative,kingma2013autoVAE}. Collectively, these theoretical advancements have not only deepened our understanding of the mechanisms behind deep generative models but also guided the development of more refined and capable generative architectures. 
    
    However, the theoretical investigation of the manifold hypothesis in deep generative models, particularly its implications in approximation theory, remains less developed. The manifold hypothesis, widely accepted as a crucial reason behind the success of these models, posits that real-world high-dimensional data often reside on lower-dimensional manifolds~\citep{bronstein2017geometric}. This concept is critical because it suggests a fundamental reason why models such as GANs and VAEs, which operate in low-dimensional spaces, are able to capture complex data distributions effectively. Although empirical evidence supports this hypothesis, a deeper theoretical understanding of how these models approximate and learn distributions on lower-dimensional manifolds within high-dimensional spaces is crucial. Further theoretical work in this area is essential, not only to validate the manifold hypothesis but also to enhance the design and efficiency of generative models in handling complex, high-dimensional data.

    Building on the previous discussion of the manifold hypothesis, \citet{dahal2022deep} offers a noteworthy theoretical contribution in the field of deep generative models. Their work effectively addresses fundamental questions in approximation theory, demonstrating that distributions on a $d$-dimensional Riemannian manifold can be approximated by a deep generative network through the pushforward measure of an easily sampled distribution in $d+1$ dimensions, such as a uniform distribution in a cube $[0,1]^{d+1}$. Importantly, they provide an upper bound on the complexity of these models, revealing that the required number of layers and neurons is determined by both the approximation error and the manifold dimension $d$. This insight offers a direct theoretical justification for the models' abilities to learn diverse probability distributions and suggesting avenues for efficiency improvements in practical applications.

    However, the research by \citet{dahal2022deep} also brings to light some important, yet unanswered questions. Primarily, the necessity of using a $d+1$ dimensional input space to approximate a $d$ dimensional manifold warrants further investigation. This approach, integral for assembling local patches, raises the possibility of achieving similar outcomes with an input dimension that matches $d$. Furthermore, in real-world applications where $d$ is not readily known, the choice of input dimension becomes a critical decision. If the chosen dimensionality is either greater or lesser than the actual $d$, how would it impact the model's performance and efficiency? Answering these questions is crucial for advancing the development of deep generative models, potentially leading to more adaptable and efficient solutions for complex data representation.

\begin{figure}[t]
            \centering
            \includegraphics[width=.55\textwidth]{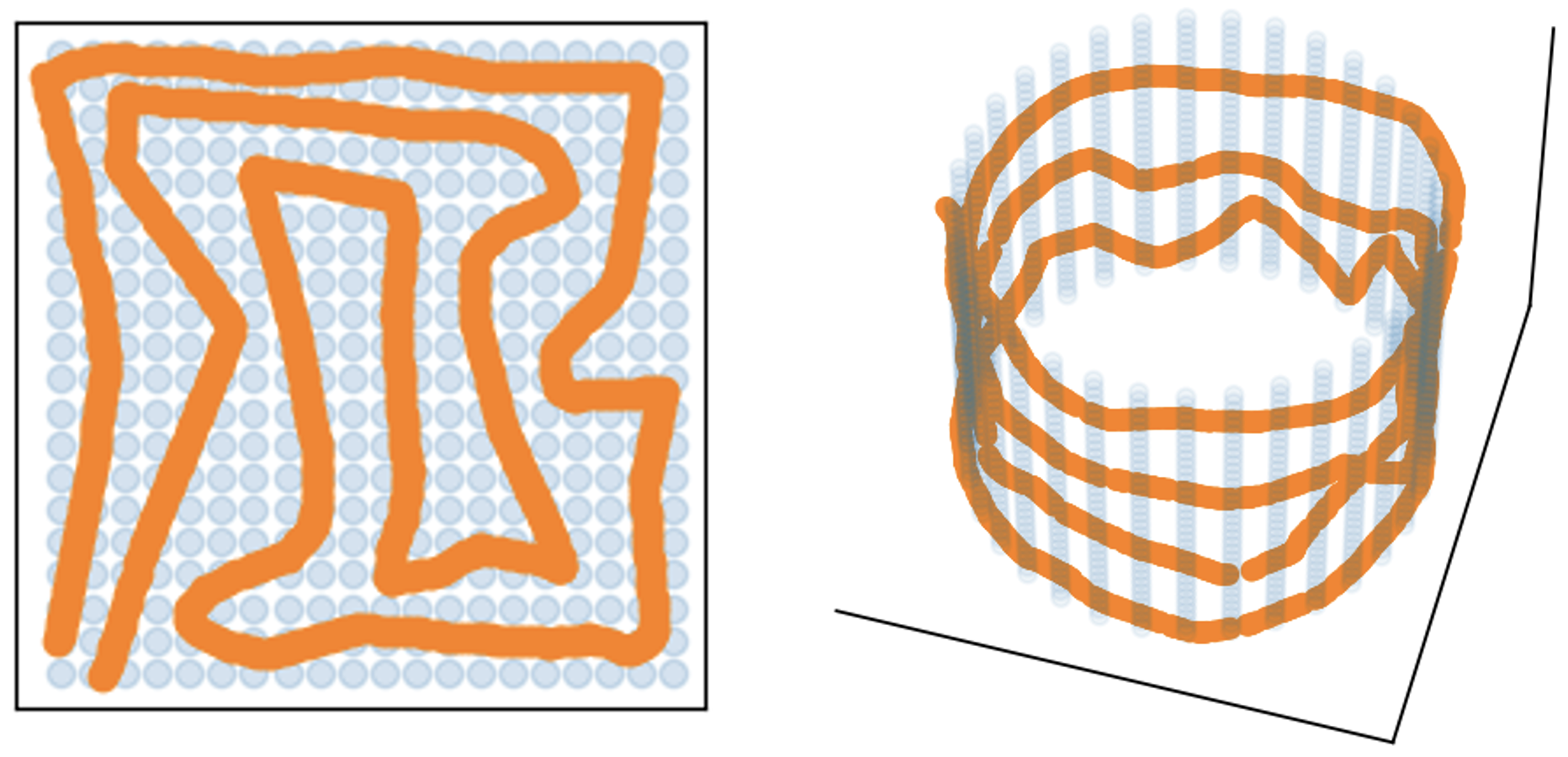}
            \caption{ \changedreviewerone{Two cases demonstrating the idea of how sufficiently large neural networks can learn distributions of higher dimension than their input sampling distributions by filling out the space. Depicted here are} 1-dimensional partial space-filling curves beginning to ``fill out" a 2-dimensional unit square (left) and a 2-dimensional cylindrical surface (right). The blue points represent the training sample \changedreviewerone{from the target distribution} used to fit the neural network, while the orange points represent new points generated by the trained network.}
            
            %\changedreviewerone{This illustrates the key idea of this paper: given sufficient size, a neural network can learn distributions of higher dimension than its input by "filling out" the space. However, the size of the network must increase massively to make up for the input dimension being too small.} }
            \label{fig:intuition}
            % \vspace{-10pt}
            \end{figure}

    In this work, we aim to understand how the input dimension relates to the manifold's intrinsic dimension in deep generative models. We adapt the space-filling curve theory to demonstrate a novel aspect of deep generative models: their ability to learn a target data distribution from easy-to-sample distributions of arbitrary dimensions. This includes the capacity to approximate these distributions even from an input as low as a one-dimensional uniform distribution on the unit interval $[0,1]$ (see \Cref{fig:intuition} for examples of curves filling 2-dimensional manifolds). \changedreviewerone{Intuitively, if the input dimension is smaller than the true dimension, the network can learn to ``fill out" the true manifold with an increasingly complex series of structures. However, in order to accurately approximate the higher dimensional structure, the approximating manifold must fold onto itself in increasingly irregular ways to fill the space.}
    
    %The key idea that we take from space-filling curve theory is that we may find curves that . For space-filling manifold approximation, the idea is analogous; manifolds which come close to touching all points of the space allow us to approximate any point in the original space by a nearby point on the approximating manifold. 
    
    %In addition, our work establishes the critical interplay among the input dimension, the true dimension of the manifold, and the approximation error. In particular, our findings highlight a ``trade-off triangle" in deep generative models, where it is unattainable to achieve low network complexity, low (underestimated) dimensionality, and low approximation error simultaneously. Specifically, when the input dimension is underestimated and the approximation error is low, we encounter one corner of this triangle: a significant increase in model complexity. This complexity, evident in the escalated number of neurons, grows super-exponentially with approximation error, underscoring the complex trade-offs and interdependencies inherent in accurately gauging the intrinsic dimension of the data. %This concept articulates a fundamental trade-off in the design of deep generative models, emphasizing the interdependent relationship between network complexity, the dimensionality of the input, and approximation error.

    In addition, our work establishes the critical interplay among the input dimension, the true dimension of the manifold, and the approximation error. \changedreviewerone{In particular, our findings highlight a ``trade-off triangle" in deep generative models, where it is unattainable to achieve low (underestimated) dimensionality, low approximation error, and low network complexity simultaneously, where network complexity is defined to be network width (number of neurons) in this paper.} Specifically, when the input dimension is underestimated and the approximation error is low, we encounter one corner of this triangle: a significant increase in model complexity. This complexity, evident in the escalated number of neurons, grows super-exponentially with approximation error, underscoring the complex trade-offs and interdependencies inherent in accurately gauging the intrinsic dimension of the data.  %This concept articulates a fundamental trade-off in the design of deep generative models, emphasizing the interdependent relationship between network complexity, the dimensionality of the input, and approximation error.
    
    This triangle metaphorically highlights the necessity of balancing these aspects when designing generative models, a central theme of our contributions. Our contribution also includes empirical studies to support these findings, which are overlooked in existing literature~ \citep{dahal2022deep}.  %Specifically, our findings suggest that simultaneously achieving low network complexity, low input dimension, and minimal approximation error is unattainable. %For instance, underestimating the input dimension necessitates a significant increase in network complexity to maintain a low approximation error. 

    Notably, our proof techniques for the underestimated dimension case are novel, requiring in-depth exploration of mathematical literature in Monge-Amp\`ere equations from the 1990s~\citep{caffarelli1990interior,caffarelli1990localization,caffarelli1991some,caffarelli1992boundary,caffarelli1992regularity,caffarelli1996boundary}, beyond the scope of existing literature~\citep{dahal2022deep,villani2009optimal}. %This work not only elucidates the deep generative networks' capabilities in learning arbitrary data distributions but also expands their applications to areas where data dimensionality is crucial.

    The structure of our paper is organized as follows: Section 2 lays the groundwork with preliminaries%, encompassing generative models, optimal transport, Riemannian manifolds, and a review of existing work including \citet{dahal2022deep}
    . Section 3 discusses our main theoretical contributions. In Section 4, we present simulations on toy cases for illustrative and visualization purposes. Section 5 offers a sketch of our proof, focusing primarily on the workflow, and is followed by a Discussion Section. Complete proofs and additional experimental details are provided in the Appendix.% which is located in the Appendix.
%\newpage

% yw / above, you may consider remove  this paragraph that describ

\section{Preliminaries}
This section covers the essential concepts underpinning our research, including generative models, optimal transport, Riemannian manifolds, and existing approximation theory for generative models on Riemannian manifolds. This groundwork is crucial for appreciating the theoretical advances we present in later sections and situates our research within the broader context of generative modeling.

\subsection{Generative Models}
    Generative models aim to learn the distribution of a dataset through sampling procedures. We consider our data following a distribution $Q$ residing in an ambient space $\mathbb{R}^D$ and an input distribution $\rho$ over a simpler space $\mathcal{S}$, such as the uniform distribution on the hypercube $[0,1]^d$. The goal is to construct a generator $g:\mathcal{S}\to \RR^D $ that minimizes a discrepancy function between the generated (pushforward) distribution $g_{\sharp }(\rho)$ and the target distribution $Q$, expressed as
    \begin{equation*}\label{eqn:discrepancy}
    \min_{g \in \mathscr{G}} \mathrm{Discrepancy}(g_{\sharp }(\rho), Q).
    \end{equation*}
    In this context, the function class $\mathscr{G}$ often comprises deep neural networks. A lower discrepancy signifies a closer match, meaning that the data generated from the simpler input space $\mathcal{S}$ closely resemble the true data represented by distribution $Q$. The essence of generative models is to produce outputs that are ``similar" to real data, aligning the generated samples with the actual distribution. A larger class $\mathscr{G}$ usually yields a smaller discrepancy, and therefore, better performance. However, this enhancement comes at the cost of increased computational resources. As a result, understanding the balance between the size of $\mathscr{G}$ and the discrepancy is of the utmost importance.%a key aspect of generative model development. 

    A practical challenge is that $Q$ is often unknown, but instead, we only observe samples $x_1,\cdots,x_n\sim Q$ that are commonly believed to be independent and identically distributed (iid) following $Q$. As a result, we can replace $Q$ by its empirical distribution $Q_n\coloneqq \frac{1}{n}\sum_{i=1}^n \delta_{x_i}$ where $\delta_{x}$ is the Dirac measure at $x$. The practical objective of generative modeling is to minimize the discrepancy between the learned distribution and this empirical distribution:
    \begin{equation*}\label{eqn:empirical_discrepancy}
    \min_{g \in \mathscr{G}} \mathrm{Discrepancy}(g_{\sharp }(\rho), Q_n),
    \end{equation*}
    where the minimizer is usually denoted by $\widehat{g}_n$.

\subsection{Optimal Transport}
In the study of generative models, the choice of discrepancy function is of fundamental importance. The Wasserstein distance, a central concept in optimal transport theory~\citep{villani2009optimal}, is commonly employed due to its effectiveness in measuring the ``cost" of transforming one distribution into another. This aligns well with the objectives of generative modeling.
\begin{definition}
    The Wasserstein-$p$ distance between two distributions $Q $ and $\nu$ in domain $\MM$, is defined as
    \[W_p(Q ,\nu) = \left(\inf_{\gamma \in \Gamma(Q , \nu)} \EE_{(x,y)\sim\gamma}(c(x,y)^p)\right)^{\frac{1}{p}},\]
    where $\Gamma(Q ,\nu)$ is the set of all couplings of $Q $ and $\nu$, containing all joint distributions over $\MM\times\MM$ with marginals $Q $ and $\nu$, and $c:\MM\times\MM\to\RR_{\geq 0}$ is called the cost function. 
\end{definition}
We focus on the Wasserstein-1 distance with $c(x,y)=\|x-y\|$, also known as the earth mover's distance. This specific case is especially relevant for our work in generative models, offering a robust framework for evaluating the similarity between generated and target distributions. \changedreviewertwo{Wasserstein-1 distance additionally admits a dual form that makes is particularly suitable for computation, which is a property that Wasserstein-$p$ in general lacks. However, our theory on convergence is extendable to Wasserstein-$p$ distance in general, which we investigate in Corollary \ref{corollary:wasserstein-p}.}

\subsection{Space-filling Curves}

In this paper, for a set $S \subset\RR^D$, we define a space-filling curve to be a continuous curve $\MM_\epsilon \subset \RR^D$ such that every point in $S$ is within $\epsilon$ of some point in $\MM_\epsilon.$ Consequently, we can define a space-filling manifold to be a manifold in $\RR^D$ satisfying the same properties:

\begin{definition}
\changedreviewertwo{A manifold $\MM_\epsilon \subset \RR^D$  is said to be $\epsilon$-space-filling manifold of $S \subset \RR^D$ if }

\[d(S,\MM_\epsilon) = \adjustlimits\sup_{x\in S} \inf_{y\in \MM_\epsilon} \|x-y\|< \epsilon.\]

$\MM_\epsilon$ is called a space-filling curve if it's a one-dimensional manifold.
\end{definition}

\changedreviewertwo{When $S$ itself is a manifold, the existence of such $\epsilon$-space-filling manifolds is shown in Lemma \ref{lem:space_filling}}. We note that this definition of the space-filling curve/manifold differs from the standard definition in topology that demands the curve be surjective on the larger space $S.$ These classical space-filling curves are typically taken as the limit of certain classes of curves:
\[\MM_{\text{space-filling}} = \lim_{\epsilon_i \to 0} \MM_{\epsilon_i},\]
where the sequence $\MM_{\epsilon_i}$ is carefully chosen to preserve desired properties (such as non self-intersection) in the limit. However, for our purposes, we do not need the limit, only particular choices of $\MM_\epsilon$ satisfying $d(S,\MM_\epsilon) < \epsilon$, and define space-filling manifolds thusly. We use the term ``space-filling" for intuition purposes, and do not demand a completely space-filling property from our manifolds.

\subsection{Riemannian Manifold}
Although data often reside in high-dimensional space $\RR^D$, there is substantial evidence suggesting that they lie on some low-dimensional manifolds~\citep{bronstein2017geometric}. This concept underlies many manifold-based generative models, such as VAEs and GANs. 

In the same manner as most existing work in the literature, we assume that the data are distributed on a $d$-dimensional orientable compact Riemannian manifold $\MM$, isometrically embedded in the ambient space $\RR^D$, with Riemannian metric $g$. A manifold is a locally Euclidean space, with each local neighbor, known as a local chart, diffeomorphic to Euclidean space $\RR^d$~\citep{boothby1986introduction}. The Riemannian metric $g$ defines a smoothly varying inner product (metric) in each tangent space $T_x\MM$. The geodesic distance $d_\MM(x,y)$ is defined as the length of the shortest path connecting $x,y\in\MM$. In addition, there exists a well-defined $d$-form, known as the Riemannian volume form $\mathrm{d}\vol_\MM$, which often serves as the analog of the Lebesgue measure on Riemannian manifold. This allows us to define density functions of probability distributions with respect to the volume measure. For more details, see \citet{do1992riemannian}. 

\subsection{Existing Work}
In previous research on generative models, there has been a focus on low-dimensional data structures, assuming that high-dimensional data are parametrized by low-dimensional latent parameters. This approach treats manifolds as globally homeomorphic to Euclidean space, implying a single-chart manifold model~\citep{luise2020generalization,schreuder2021statistical,block2022intrinsic,chae2023likelihood}. However, this assumption presents limitations in dealing with the complexity inherent in general manifolds with multiple charts. 

In contrast, \citet{yang2022capacity} and \citet{huang2022error} demonstrate that GANs can approximate any data distribution from a one-dimensional continuous distribution. This method does not assume a global chart; however, it heavily relies on GANs memorizing empirical data distributions, posing limitations in generating novel samples.

To overcome the limitations of the above approaches, \citet{dahal2022deep} represents a significant development in this context. By avoiding the single-chart assumption and the need for memorizing data, they construct an oracle transport map suitable for general manifolds with multiple charts. This advancement in approximating distributions on Riemannian manifolds with neural network pushforwards paves the way for more sophisticated and effective generative models. Their study relies on two mild assumptions:
\begin{assumption}\label{as:1}
 $\MM$ is a $d$-dimensional compact Riemannian manifold isometrically embedded in $\RR^D$. As a consequence, there exists $B>0$ such that $\|x\|_\infty\leq B, \forall x\in\MM$. 
\end{assumption}
\begin{assumption}\label{as:2}
$Q$ is supported on $\MM$ and has a density function $q$ with respect to the volume measure $\vol_\MM$ with a positive lower bound $c$: $0<c\leq q(x),\forall x\in\MM$. 
\end{assumption}
We consider the ReLU-type neural networks
\begin{equation*}\label{eqn:relu}
g(x)=W_{L}\sigma(W_{L-1}\cdots\sigma(W_{1}x+b_{1})+\cdots+b_{L-1})+b_{L},
\end{equation*}
where $\sigma$ is the ReLU activation function~\citep{fukushima1969visual}, $W$ is the weight matrix, $b$ is the bias vector. The function class, denoted by $\mathscr{G}_{NN}(m,L,p,\kappa)$, contains neural networks with ReLU activation function, input dimension $m$, maximum depth $L$, maximum width $p$, and bounded weights and biases by $\kappa$:
   \begin{align*}
        &\mathscr{G}_{NN}(m,L,p,\kappa)\coloneqq \left\{g=(g_1,\cdots,g_D):\RR^{m}\to \RR^D:\right.\\
        &~~~~~~~~ g_j\text{ is in form \eqref{eqn:relu} with at most $L$ layers and max width $p$}, \left. \|W_{i}\|_\infty\leq \kappa, \|b_{i}\|_\infty\leq \kappa\right\}.
    \end{align*}

Under the above assumptions, the two main theorems in \citet{dahal2022deep} are as follows:

Firstly, Lemma \ref{lem:Dahal1} establishes that certain distributions on compact Riemannian manifolds can be approximated using a deep neural network and a simple input distribution. Crucially, the dimension of the input space is precisely one more than the manifold dimension. Furthermore, Lemma \ref{lem:Dahal1} provides detailed complexity bounds on the size of the network to achieve a specified level of accuracy measured by the Wasserstein loss.

\begin{lemma}[Theorem 1 of \citealt{dahal2022deep}]\label{lem:Dahal1}
Let $\rho=\rm{Unif}(0,1)^{d+1}$, then there exists a constant $0<\alpha<1$ that is independent of $D$ such that for any $0<\epsilon<1$, there exists a deep neural network $g\in \mathscr{G}_{NN}(d+1,L,p,\kappa)$ with $L=O\left(\log\left(\frac{1}{\epsilon}\right)\right)$, $p=O\left(D\epsilon^{-\frac{d}{\alpha}}\right)$, $\kappa =B$, that satisfies $W_1(g_\sharp(\rho),Q)<\epsilon$. 
\end{lemma}

Next, Lemma \ref{lem:Dahal2} serves as the empirical counterpart of Lemma \ref{lem:Dahal1}, addressing scenarios in which only finite samples following the distribution $Q$ are observed. It demonstrates the theoretical possibility of identifying such generators using finite samples instead of direct access to the true distribution $Q$. In addition, this lemma links the network size to the sample size rather than the approximation error.

 \begin{lemma}[Theorem 2 of \citealt{dahal2022deep}]\label{lem:Dahal2}
 Under the same assumption as in Lemma \ref{lem:Dahal1}, let $x_1,\cdots,x_n\stackrel{iid}{\sim} Q$ be $n$ iid samples from $Q$, then for any $\delta>0$, set $\epsilon=n^{-\frac{1}{d+\delta}}$ in Lemma \ref{lem:Dahal1} so that the network class $\mathscr{G}_{NN}(d+1,L,p,\kappa)$ has parameters $L=O\left(\log\left(n^{\frac{1}{d+\delta}}\right)\right)$, $p=O\left(Dn^{\frac{d}{\alpha(d+\delta)}}\right)$, $\kappa =B$. Then the empirical risk minimizer $\widehat{g}_n$ has rate $\EE\left[W_1(\widehat{g}_{n\sharp}(\rho),Q)\right]\leq Cn^{-\frac{1}{d+\delta}}$, where $C$ is a constant independent of $n$ and $D$. 
\end{lemma}   
 These results shed light on how distributions on the manifold can be approximated by a deep neural network's pushforward of a low-dimensional easy-to-sample distribution. 

 However, this approach has several limitations. First, the intrinsic dimension $d$ of the manifold $\MM$ is almost never known. Although there is an immense literature on estimating $d$~\citep{levina2004maximum,zheng2022learning,horvat2022intrinsic,brown2022union}, it has been proven to be an almost insurmountable problem due to its complexity~\citep{fefferman2016testing}. Second, even if $d$ is known, the choice of input dimension $d+1$ is unnatural, as it exceeds the true dimension. Technically, the extra dimension results from the need to connect local neighborhoods using a pasting algorithm. But this seems superfluous from an intrinsic perspective, considering partition of unity could potentially serve the same purpose. Third, in practice, the input dimension is often treated as a tuning parameter, or simply set based on historical experience or recommendations without rigorous optimization. It remains unclear how these decisions on input dimension affect approximation performance, especially when $d$ is underestimated. 
 
To address these issues, our study examines the scenario where the input dimension is arbitrary, potentially as minimal as one, within the same framework. We will summarize our primary findings in the subsequent section.

\section{Main Theory}
    We begin with \citet{dahal2022deep}, which presents a method for approximating distributions on compact Riemannian manifolds using neural networks with an input dimension of $d+1$. This raises a natural question: \emph{can the input dimension be reduced to 
$d$, the manifold's intrinsic dimension, or even lower, without compromising the approximation's effectiveness?} In this work, we answer this question affirmatively, leveraging concepts of space-filling curves and insights from a series of work on Monge-Amp\`ere equations by Caffarelli~\citep{caffarelli1990interior,caffarelli1990localization,caffarelli1991some,caffarelli1992boundary,caffarelli1992regularity,caffarelli1996boundary}. Our theorem demonstrates the feasibility of using any input dimension $m\geq 1$, which can be either smaller or larger than $d$, even down to 1. However, reducing the input dimension below the manifold's dimension introduces a super-exponential increase in complexity. Finally, we highlight that our approach adheres to the same foundational assumptions as \citet{dahal2022deep}.    

% Acknowledging the variability in input dimension size 
% $m$, which was treated as a constant in existing literature, we adjust the definition of the neural network function class to emphasize its dependency on $m$. This class, denoted as $\mathscr{G}_{NN}(m,L,p,\kappa)$, allows for a more targeted exploration of the networks' capabilities across different input dimensions：
%      \begin{align*}
%         &\mathscr{G}_{NN}(m,L,p,\kappa)\\
%         &\coloneqq \left\{g=(g_1,\cdots,g_D):\RR^{m}\to \RR^D:\right.\\
%         & g_j\text{ is in form \eqref{eqn:relu} with at most $L$ layers and max width $p$},\\
%         & \left. \|W_{i}\|_\infty\leq \kappa, \|b_{i}\|_\infty\leq \kappa\right\}.
%     \end{align*}
    Our theorem, central to this discussion, asserts the approximation power of these neural networks under varying input dimensions. It establishes the conditions under which the networks can effectively approximate distributions on Riemannian manifolds, taking into account the dimensionality of the input and its impact on the network's complexity.
    
    \begin{theorem}[Approximation Power of Deep Generative Models]\label{thm:main_population} 
    Under Assumptions \ref{as:1} and \ref{as:2}, and with $\rho$ as the uniform measure on $[0,1]^m$, for any $\epsilon>0$, there exists a deep neural network $g\in \mathscr{G}_{NN}(m,L,p,\kappa)$ such that
    $L=O\left(\log\left(\frac{1}{\epsilon}\right)\right)$, $p=\begin{cases}O\left(D\epsilon^\frac{-m}{\alpha(m,\epsilon)}\right) & m\leq d \\
    O\left(D\epsilon^\frac{-d}{\alpha}\right) & m> d\end{cases}$, $\kappa = \max\{B,1\}$, and $W_1(g_\sharp(\rho),Q  )<\epsilon$. Furthermore, when $m\leq d$, $\lim_{\epsilon \to 0} \alpha(m,\epsilon) = 0$, leading to a super-exponential increase in the width $p$. 
        \end{theorem}

    The purpose of this theorem is two-fold. Firstly, it demonstrates that neural networks can approximate distributions on a manifold, with the flexibility to use an input distribution uniformly distributed over a hypercube $[0,1]^m$ of any dimension. This means that the choice of $m$ can be adapted as needed, whether it's smaller or larger than the manifold dimension $d$. We additionally note that the choice of our generator to be the uniform distribution on the unit hypercube is non-restrictive. These results can easily be extended to any input distribution with a density that is upper bounded away from infinity and lower bounded from zero on the unit cube, such as a truncated normal distribution or uniform distributions.

    Secondly, the theorem provides a quantitative link between the complexity of the network and the relationship between the input dimension $m$ and the manifold dimension $d$. Notably, when $m> d$, the rates coincide with those found in \citet{dahal2022deep}. However, the situation becomes particularly intriguing when the input dimension $m$ is smaller than the target dimension $d$. In this scenario, the width of the neural network no longer increases at a polynomial rate of $\frac{d}{\alpha}$, but at a super-exponential rate of $\frac{m}{\alpha(m,\epsilon)}$ where $\frac{m}{\alpha(m,\epsilon)} \xrightarrow[]{\epsilon\to0} \infty.$ This phenomenon implies that, while it is theoretically feasible to approximate a broad class of distributions with low-dimensional inputs, choosing an appropriate input dimension is critical to avoid excessively complex networks; it is often more advantageous to overestimate $d$ than to underestimate it.
    
%Additionally, a pivotal aspect of our proof is noteworthy: it demonstrates the existence of a manifold with arbitrary dimension, potentially down to one, that effectively ``fills" up the target manifold. We further show the feasibility of learning a distribution on this ``space-filling manifold" using a deep neural network. Practical visual examples of this phenomenon are detailed in \Cref{sec4}.

    We now turn our attention to the empirical aspect of our theory, examining the statistical guarantees provided by our model in the presence of iid samples $x_1,\cdots,x_n\sim Q$. 

    \begin{theorem}[Statistical Guarantees of Deep Generative Models]\label{thm:main_empirical}
        Given iid samples $x_1,\cdots,x_n\sim Q$ with empirical distribution $Q_n $, let $\delta>0$, $L=O\left(\log\left(n^{\frac{1}{d+\delta}}\right)\right)$, 
        $$p=\begin{cases}O\left(Dn^{\frac{m}{(m+\delta)\alpha(m,n,d,\delta)}}\right) & m \leq d\\
        O\left(Dn^{\frac{d}{(d+\delta)\alpha}}\right)& m > d,\end{cases}$$ $\kappa = \max\{B,1\}$, then the empirical risk minimizer $\widehat{g}_n\in\mathscr{G}_{NN}(m,L,p,\kappa)$ satisfies 
        \[\EE \left [ W_1(\widehat{g}_{n\sharp }(\rho),Q) \right ] \leq (1+2C_\delta)n^{-\frac{1}{d+\delta}},\]
        where $C_\delta$ is a constant, independent of $n$. Furthermore, when $m\leq d$, $\lim_{n \to \infty} \alpha(m,n,d,\delta) = 0$, leading to a super-exponential increase in the width $p$. 
    \end{theorem}

Similar to the population version, this theorem demonstrates that deep neural networks can approximate manifold distributions from 
$n$ iid observations, with the approximation error diminishing at a rate of $\frac{1}{d+\delta}$. However, the complexity of the network, particularly the width, increases alongside $n$. Although a larger $n$ yields a smaller approximation error, when $m\leq d$, the width $p$ grows super-exponentially with $n$. 

These two theorems collectively underscore a ``trade-off triangle'' of model complexity, input dimension, and approximation accuracy. In the next section, we illustrate these concepts through simulations using $m=1,2, 3$ as case studies for proof of concept and visualization.

\changedreviewertwo{Finally, we also show that \Cref{thm:main_population} and \Cref{thm:main_empirical} can be partially extended to Wasserstein-$p$ distances in the following corollary.} 

\begin{corollary}\label{corollary:wasserstein-p}
    \changedreviewertwo{Under the same setup as \Cref{thm:main_population}, for any $\epsilon>0$, there exists a deep neural network $g\in \mathscr{G}_{NN}(m,L,p,\kappa)$ such that $W_p(g_\sharp(\rho),Q  )<\epsilon$. Moreover, under the same setup as \Cref{thm:main_empirical}, the empirical risk minimizer $\widehat{g}_n\in\mathscr{G}_{NN}(m,L,p,\kappa)$ satisfies $\EE \left [ W_p(\widehat{g}_{n\sharp }(\rho),Q) \right ] \leq (1+2C_\delta)n^{-\frac{1}{d+\delta}}$.}
\end{corollary}
\changedreviewertwo{Note that the bounds for $L,p,\kappa$ are not provided in this case; a more detailed discussion is presented in the proof in \Cref{apdx:wasserstein_p}.}

\section{Simulation}\label{sec4}

    In this section, we provide empirical evidence on three toy examples for visualization purposes of these deep generative networks' abilities to learn space-filling curve approximations. We demonstrate the ability of the standard ReLU network to map the uniform distribution on the unit hypercubes of various dimensions $[0,1]^m$ to a variety of target distributions, where $m$ is the dimension of the input distribution. Note that here we are using a standard feedforward neural network with a Wasserstein-1 loss function approximation provided in the Python Optimal Transport package (POT, \citealp{flamary2021pot}). This loss function takes the role of the ``critic" network in a typical GAN model, hence our fitted feedforward network under this Wasserstein-1 loss function is analogous to the ``generator" in a Wasserstein GAN \citep{arjovsky2017wasserstein}. We also report an empirical ``fill distance" \changedreviewerone{which we measure by taking a set of samples $x_1,...,x_n$ from the true manifold, another set of  samples $y_1,...,y_N$ from the generated manifold, and calculating $\frac{1}{n}\sum_{i=1}^{n} \min_{y_j} \left\{d(x_i,y_j)\right\}$. This approximates the mean distance over all points in the data manifold to the closest point on the approximating curve.} Hence, this is a measurement of the distance between the manifold supports of distributions, while the Wasserstein loss (used to train the neural networks) offers a measurement of the distance between distributions.

    In Simulation 1, we show that the problem of finding a generator from the two dimensional square to the two dimensional square is computationally feasible (as in the required complexity is relatively small). Additionally, we show that the same problem with only a one dimensional input dimension is relatively harder, and also show the trajectory of training iterations to visualize how the neural network ``learns" the distribution with an under-dimensioned input. In Simulation 2, we showcase the mapping of the same uniform distribution on $[0,1]$ to the uniform distribution on a 2-dimensional cylinder $S^2\subset\RR^3$ ($m=1,d=2,D=3$). We also showcase that the same target distribution can be learned with much lower complexity when we match the input dimension to the manifold dimension, that is $m=2=d$. In Simulation 3, we take our target distribution to be the 3-dimensional uniform distribution on a unit cube $[0,1]^3$~($d=D=3$). As in Simulations 1 and 2, we train ReLU networks using uniform distributions on the 1, 2, and 3 dimensional hypercubes ($m=1,2, 3$) and compare performance. Full implementation details and additional simulations \changedreviewerone{including a study of network complexity and approximation error, the use of higher input dimension than target manifold dimension ($m=d+1$) for each of the three simulation cases, and the case of non-uniform input or target distributions}, along with a link to a GitHub repository containing all code, can be found in \Cref{apdx:exp}.

             \begin{figure}[h!]
            \centering
            \includegraphics[width=.7\textwidth]{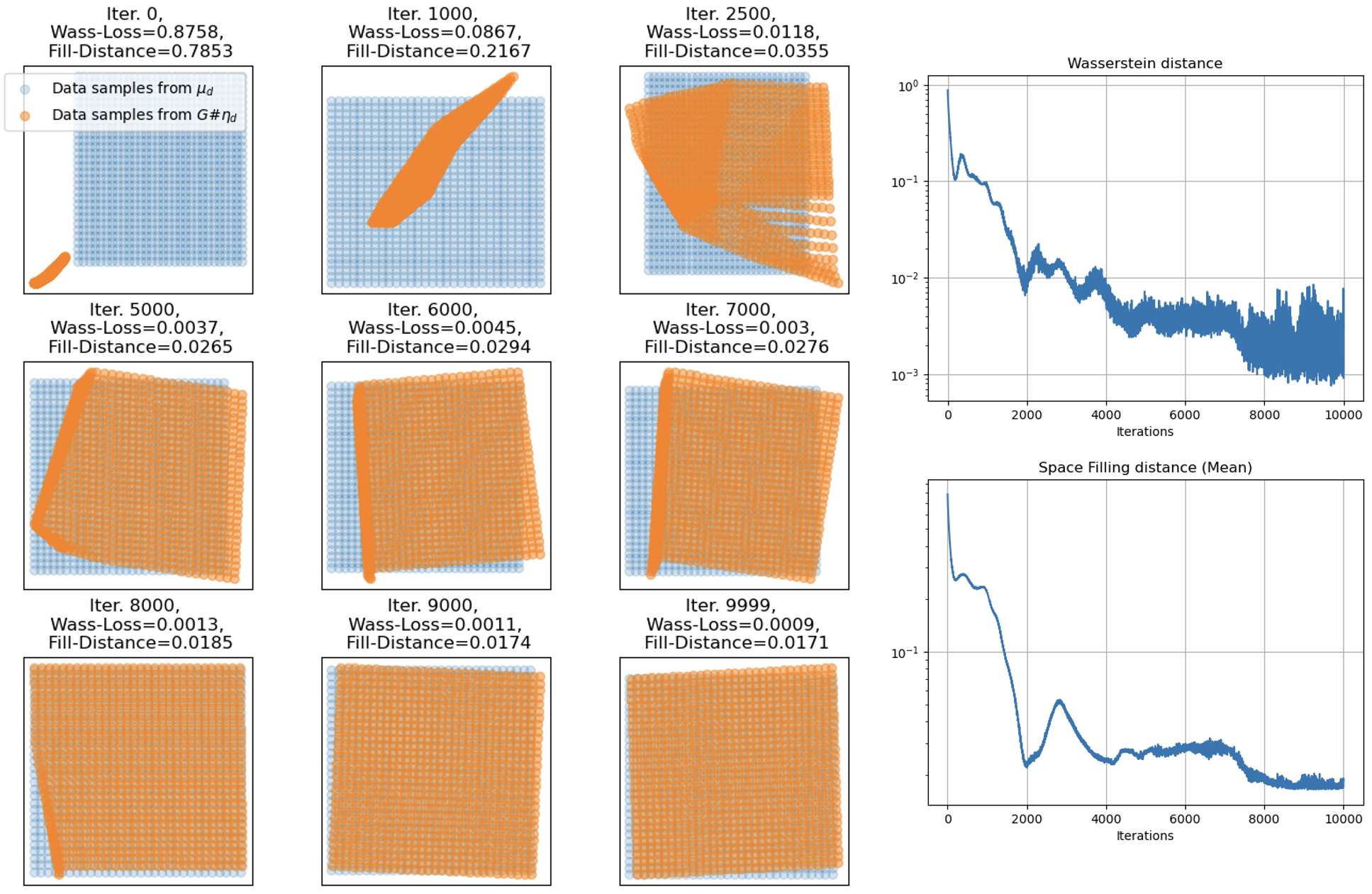}
            \caption{Simulation 1: Training results (left) of a small 2 hidden layer, 10 node each, fully connected network mapping the 2-D uniform input to a 2-D uniform target data manifold. The orange surface is generated by the neural network and is ``filling in'' the data manifold (blue). Wasserstein loss (top right) and fill distance (bottom right) between generated and observed data per training iteration for a total of 10,000 iterations. }
            \label{fig:sim1-2d2d-all}
        \end{figure}

    \subsection{Simulation 1}\label{sec:sim1}
        In Simulation 1, we let the data distribution $Q$ be a uniform distribution over the unit square $[0,1]^2$ and consider the cases where the input distribution $\rho$ to the neural network is a uniform distribution over the unit square $[0,1]^2 \in \RR^2$ and a uniform distribution over the unit interval $[0,1] \in \RR$.  In this case, the manifold dimension $d=2$ is equal to the dimension of the ambient space $D=2$. We first consider the case where the input and target dimensions are equal.

        \begin{figure}[h!]
            \centering
            \includegraphics[width=.7\textwidth]{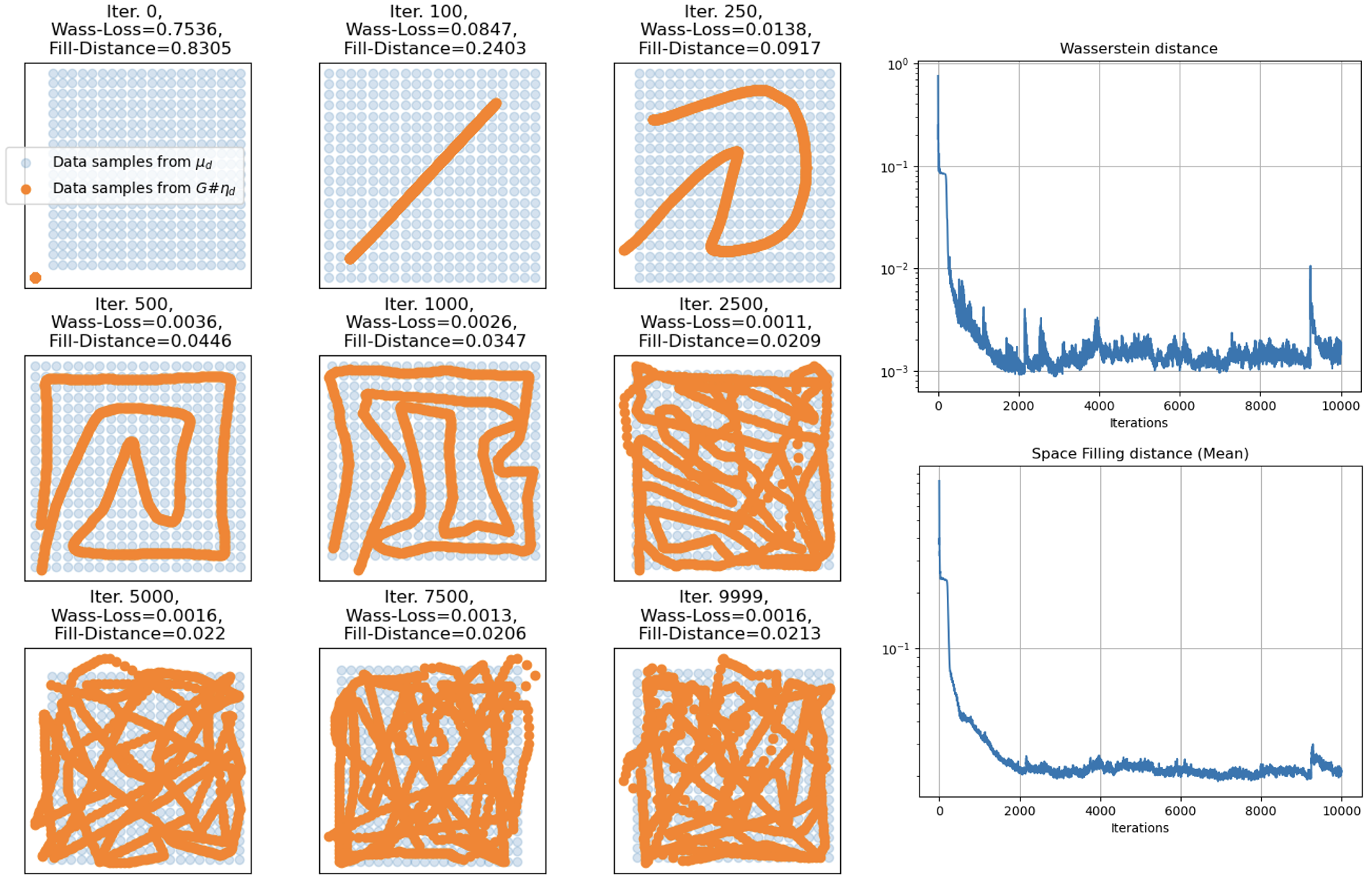}
            \caption{Simulation 1: Training results of a 5 hidden layer, 200 node each, fully connected network for 10,000 iterations. Training trajectory (left), Wasserstein loss (top right), and fill distance (bottom right) of mapping a uniform distribution on the interval $[0,1]$ to a uniform distribution on the unit square $[0,1]^2$. It can be seen that as the loss decreases, the fitted curve fills more of the square as expected.}
            \label{fig:sim-1-trajectory-main}
            % \vspace{-10pt}
        \end{figure}
        
     Aligned with our theory, the required complexity in this case is expected to be significantly lower than when the target data dimension is underestimated. In Figure \ref{fig:sim1-2d2d-all}, we show that training a fully connected feed forward network with just 2 hidden layers with 10 nodes each for 10,000 iterations under Wasserstein loss, again from the Python Optimal Transport (OT) package, achieves low Wasserstein loss as well as low average empirical fill distance. Note that in this case, during training the input is taken to be a random sample on the unit square at each iteration, while the target data is a uniform grid on the unit square, so that the input sample is not identical to the target data at any iteration.

    Next we re-examine the target distribution from the case that the input dimension $\rho$ is a uniform distribution over the unit interval $[0,1]$. \Cref{fig:sim-1-trajectory-main} demonstrates how a standard feedforward, fully-connected neural network, using ReLU activation and trained with Wasserstein loss, effectively maps the unit interval into an increasingly complex, space-filling curve in a two-dimensional space. Note that now a much larger network size of 5 hidden layers with 200 nodes each is required to reach a similar level of approximation to the equal input and target dimensions case. For further simulations regarding the effect of different network sizes see \Cref{apdx:sim1}, and for the $m=d+1$ case see \Cref{apdx:sim_m=d+1}.

    \subsection{Simulation 2}

    Analogous to Simulation 1, in Simulation 2 we consider a situation where the data distribution $Q$ is a uniform distribution over a cylinder embedded in $\RR^3$ and again consider the cases where the input distribution $\rho$ to the neural network is a uniform distribution over the unit square $[0,1]^2 \in \RR^2$ and the unit interval $[0,1]\in \RR$.
    
\begin{figure}[h!]
            \centering
            \includegraphics[width=.7\textwidth]{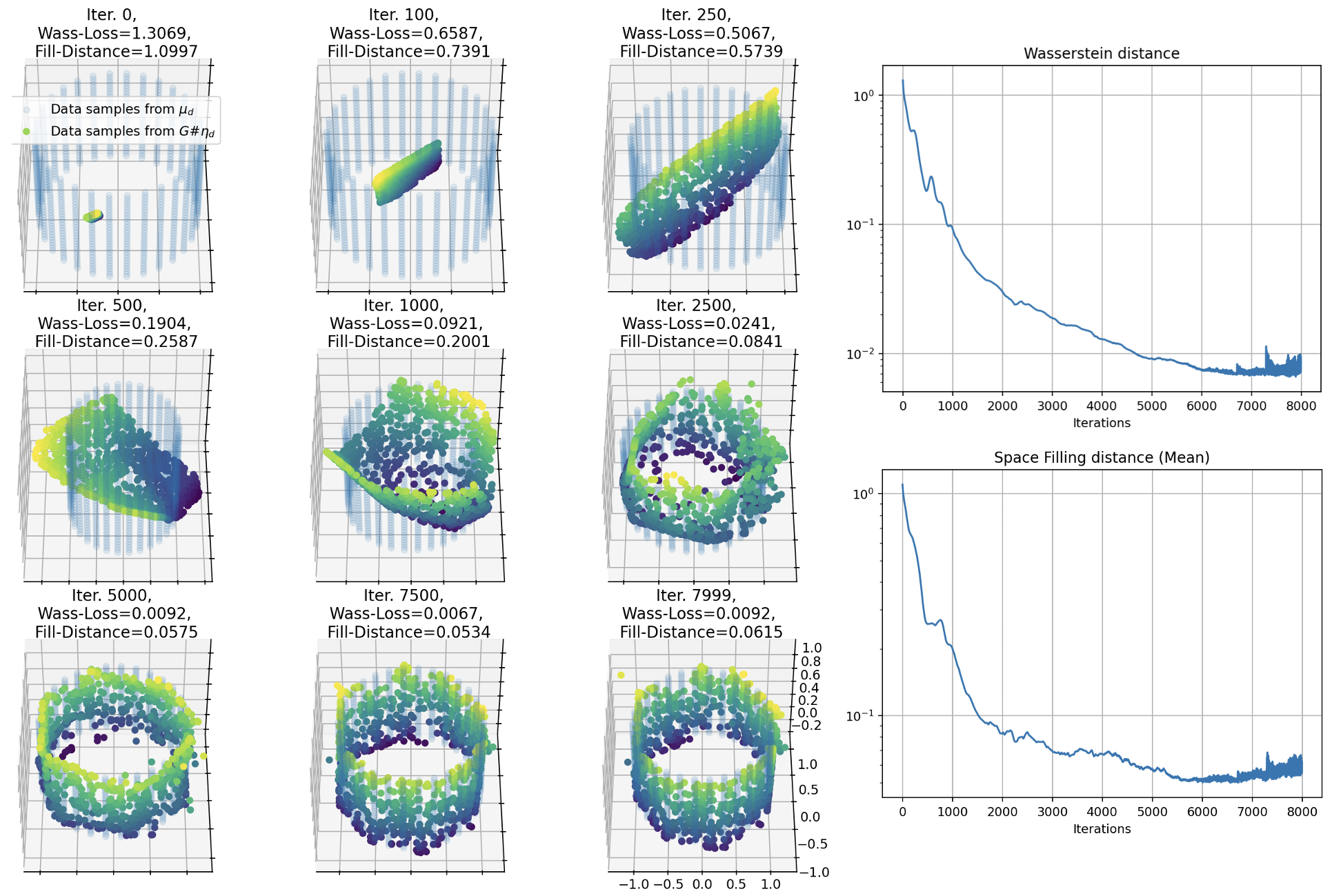}
            \caption{Simulation 2: Training results of a small 3 hidden layer, 25 node each, fully connected network for 8,000 iterations mapping a 2-D uniform input to a 2-D uniform target data manifold on a cylinder embedded in $\RR^3$. Here the multicolored surface is generated by the neural network (colored by $z$-axis height) and is ``filling in'' the data manifold (blue).}
            \label{fig:sim2-2d2d-trajectory}
            \vspace{-10pt}
            \end{figure}

    We first consider when the input distribution (uniform) dimension matches the target data manifold dimension (also uniform), where $m=d=2$. We show that, aligned with our theory, the required complexity in this case is observed to be much lower than when the target data dimension is underestimated. In Figure \ref{fig:sim2-2d2d-trajectory}, we show the results of training a fully connected feed forward network with a total of 3 hidden layers with 25 nodes each for 8,000 iterations under Wasserstein loss and that the network is able to achieve low Wasserstein loss as well as low average fill distance.  
           
\begin{figure}[h!]
           \centering
            \includegraphics[width=.7\textwidth]{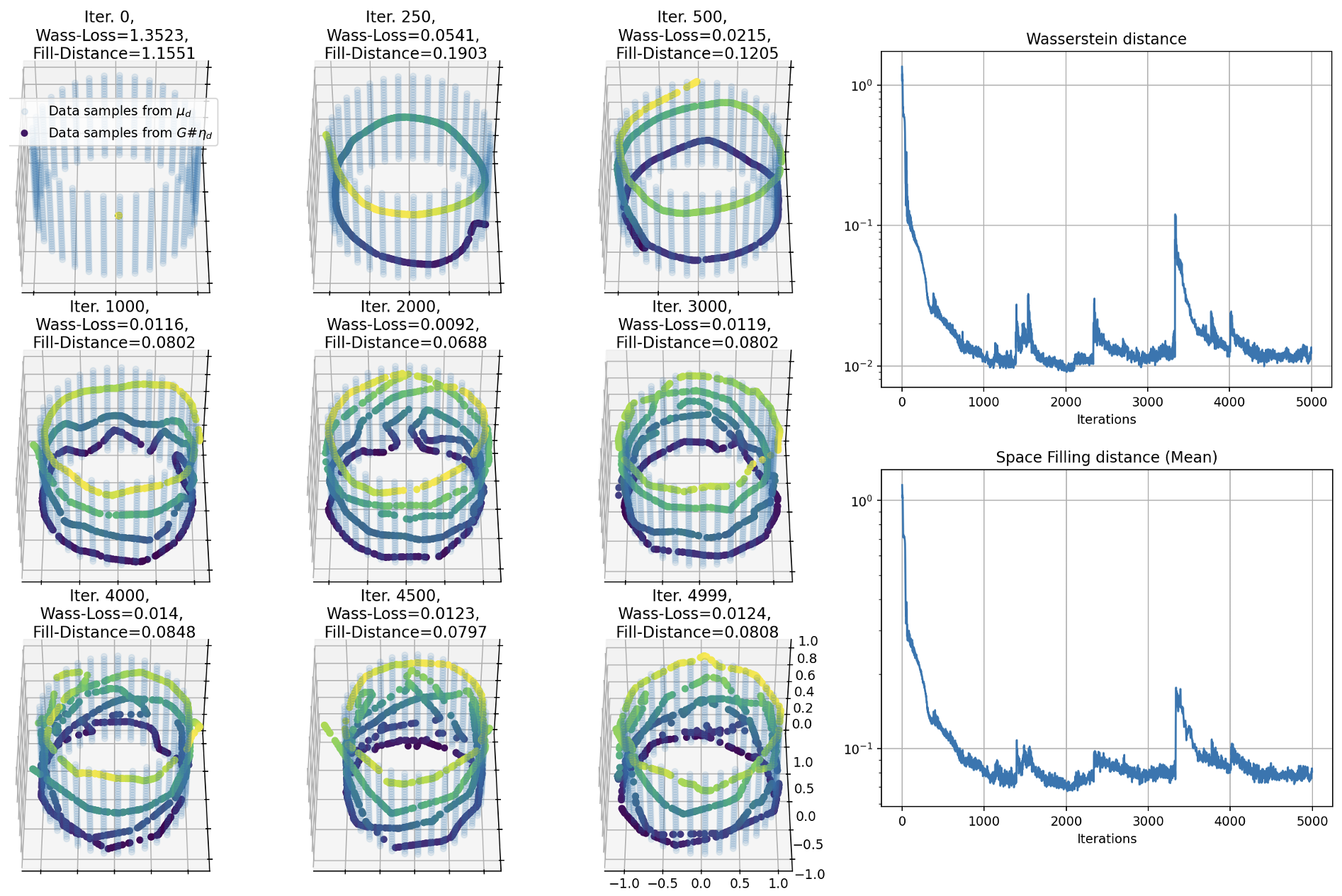}     
            \caption{Simulation 2: Training results of a 7 hidden layer, 250 node each, fully connected network over 5,000 iterations. Training trajectory (left), Wasserstein loss (top right), and fill distance (bottom right) of mapping a 1-D uniform distribution on the interval $[0,1]$ to a 2-D uniform distribution on a cylinder embedded in $\RR^3$. As the loss decreases over the iterations, the fitted curve fills more of the surface. }
            \label{fig:sim-2-training}
            % \vspace{-10pt}
            \end{figure}

    In contrast, we consider the same uniform distribution on a cylinder embedded in $\RR^3$ but with an input distribution of only one dimension. Note again that in this example we have the manifold $\MM$ with dimension $d=2$, embedded in an ambient Euclidean space with dimension $D=3$. We show again that a feedforward, fully-connected neural network under Wasserstein loss and ReLU nonlinearity is still able to learn space-filling curves properly on the manifold of lower dimension than the ambient space (albeit with a much larger network size of 7 hidden layers with 250 nodes each). Again, the neural network maps the input interval into an increasingly complex space-filling curve on the cylinder as the number of training iterations increases and the loss decreases respectively as seen in Figure \ref{fig:sim-2-training}. For full implementation details and code, as well as the $m=d+1$ case, see \Cref{apdx:exp}.

     \subsection{Simulation 3}

     In Simulation 3, we consider a uniform distribution on a unit cube $[0,1]^3$ in $\RR^3$, while now taking as input one, two, and three-dimensional uniform distributions on $[0,1]^m$ for $m=1, 2, 3$ respectively. We show again that a feedforward, fully-connected neural network under Wasserstein loss and ReLU nonlinearity is still able to learn ``space-filling sheets" on the manifold of higher dimension than the input.

      %Finally, we demonstrate that the previous results also hold when applied to Simulation 3 of the main paper. Again, we first add an experiment to the $m=d$ case for our example mapping uniform input to a uniform distribution on the unit cube to as comparison for the following $m<d$ cases. Again for this case, the input sample is redrawn each iteration while the target is a constant grid on the unit cube, hence the input and target are not identical in any training iteration. 
      
\begin{figure}[h!]
             \centering
            \includegraphics[width=.7\textwidth]{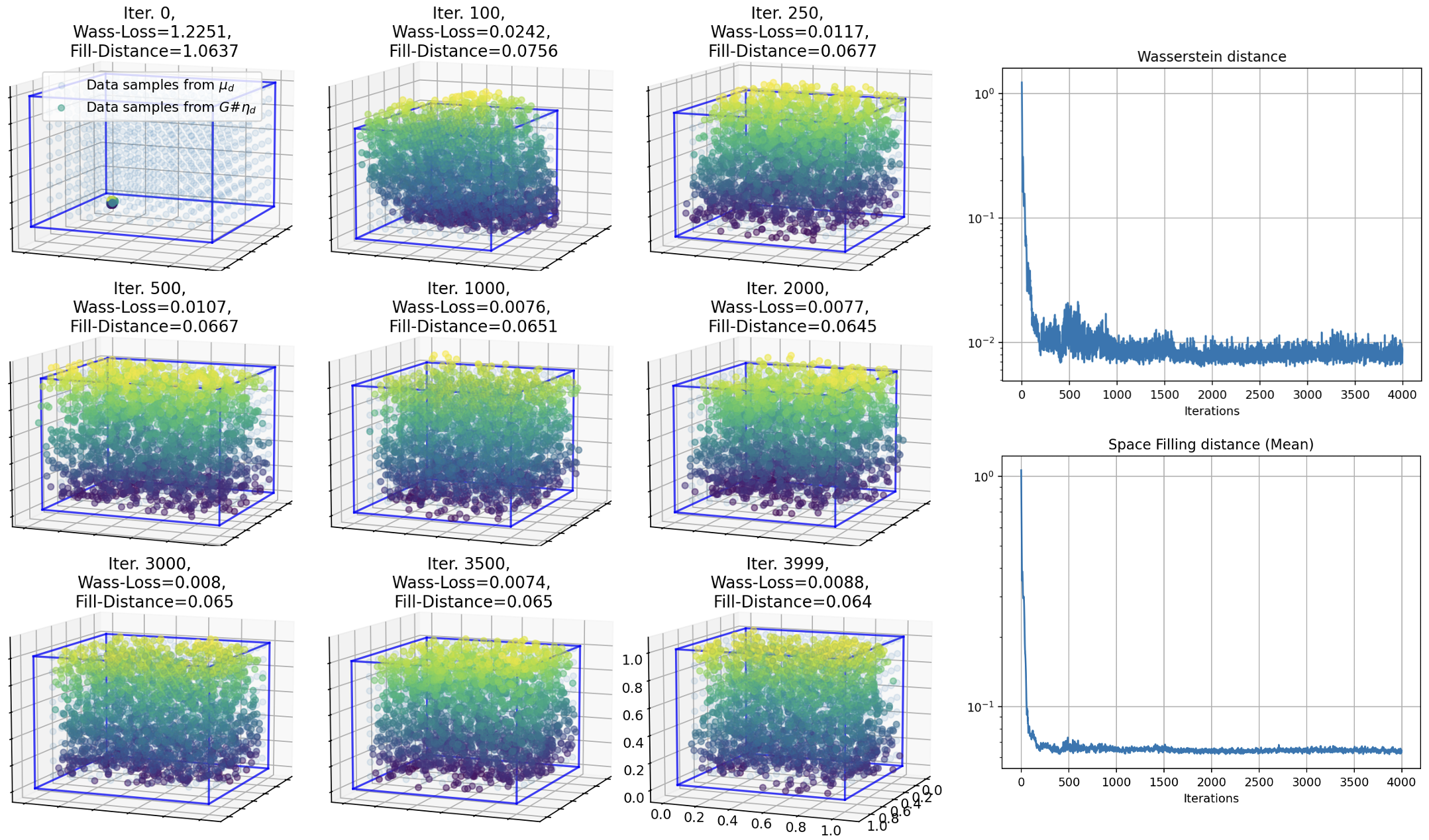}
            \caption{Simulation 3: Training results of a 3 hidden layer, 128 node each, fully connected network over 4,000 iterations mapping the 3-D uniform input to a 3-D uniform target data manifold on the unit cube. Here the multicolored curve is generated by the neural network (colored by $z$-axis height) and is ``filling in'' the data manifold (blue).}
            \label{fig:sim3-3d3d-trajectory}
            % \vspace{-10pt}
            \end{figure}

      We again start with the case where the input dimension matches the target dimension ($m=3$). In Figure \ref{fig:sim3-3d3d-trajectory},
      we show the results of training a neural network with 3 hidden layers of 128 neurons each. It can be seen that the network achieves both low Wasserstein loss as well as empirical fill distance in Figure 
      %\ref{fig:sim3-3d3d-losses}.
      \ref{fig:sim3-3d3d-trajectory}.
      Note that in this case the input data is a randomly re-sampled each training iteration, while the target remains a fixed grid on the unit cube, such that the input and target data are not identical at any iteration. 
      
        %\begin{figure}[!h]
        %    \centering
        %    \includegraphics[width=.45\textwidth]{ICML-2024/figures-rebuttal/sim3-2d3d-trajectory.png}
        %    \caption{Simulation 3: Training trajectory of mapping a uniform distribution on the unit square $[0,1]^2$ to a uniform distribution on $[0,1]^3$, colored by height (Z-axis). As the loss decreases over the iterations, the fitted surface fills more of the space in the cube. }
      %      \label{fig:sim-3-training}
      %      \vspace{-10pt}
      %      \end{figure}

\begin{figure}[h!]
            \centering
            \includegraphics[width=.7\textwidth]{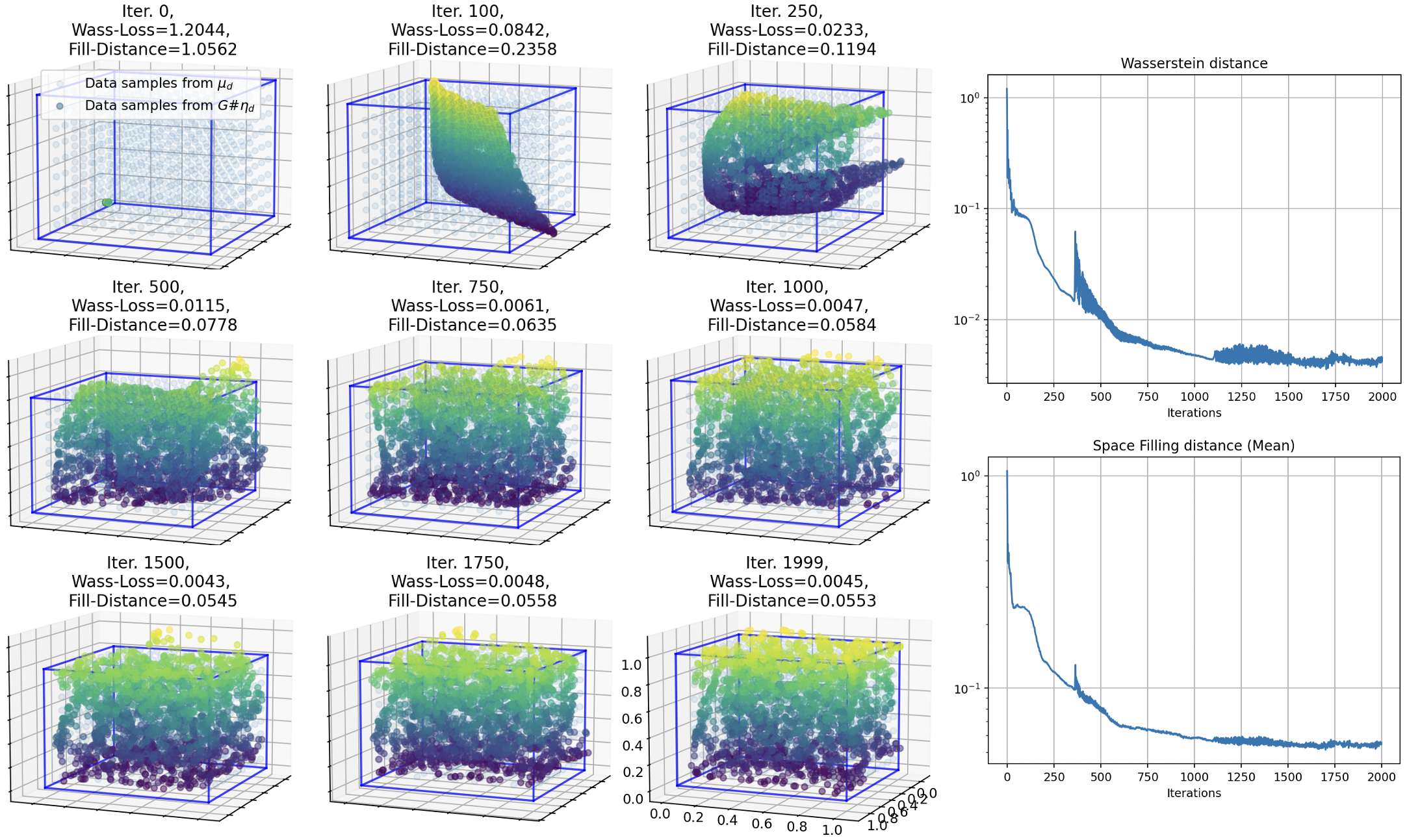}
            \caption{Simulation 3: Training results of a 4 hidden layer, 256 node each, fully connected network over 2,000 iterations mapping a uniform distribution on the unit square $[0,1]^2$ to a uniform distribution on $[0,1]^3$, colored by height ($z$-axis). As the loss decreases over the iterations, the fitted surface fills more of the space in the cube. }
            \label{fig:sim-3-training}
            % \vspace{-10pt}
            \end{figure}

             Now we reduce the input dimension by 1, mapping a uniform distribution of dimension $m=2$ to a uniform distribution in a higher target dimension $d=3$. We observe in \Cref{fig:sim-3-training} that a larger neural network with 4 hidden layers of 256 nodes each achieves similar loss to the smaller model used when $m=d=3$.

           \begin{figure}[h!]
            \centering
            \includegraphics[width=.7\textwidth]{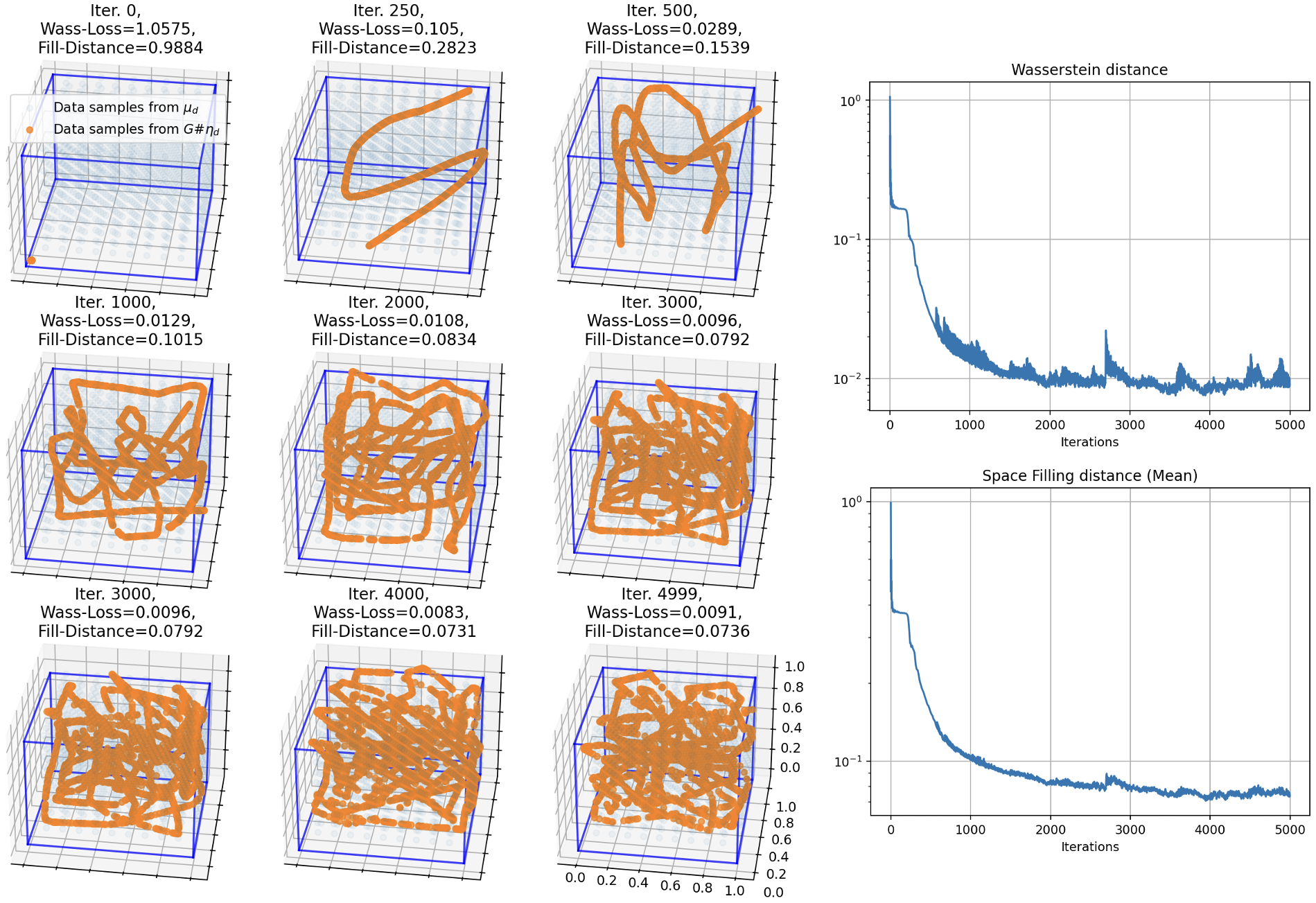}
            \caption{Simulation 3: Training results of a 5 hidden layer, 256 node each, fully connected network over 5,000 iterations mapping the 1-D uniform input on the interval [0,1] to a 3-D uniform target data manifold on the unit cube. Here the orange curve is generated by the neural network and is ``filling in'' the data manifold (blue). }
            \label{fig:sim3-1d3d-trajectory}
            % \vspace{-10pt}
            \end{figure}
            
        Finally, we reduce the input dimension one more time for the case when the input dimension is $m = 1$  and the data manifold is uniform in $d=3$ dimensions. We observe in \Cref{fig:sim3-1d3d-trajectory} that again a larger network with 5 hidden layers of 256 neurons each is required to achieve similar loss to the $m=2$ and $m=3$ cases. Together, the three simulations underscore the networks' ability to adaptively map lower-dimensional inputs onto higher-dimensional manifolds in a generalizable manner, effectively demonstrating our main theorems in practice. Again, see \Cref{apdx:exp} for full implementation details and extension to $m=d+1$. 
      % \newpage

\section{Sketch of the Proof}

In this section, we provide a proof sketch of our main theorems, highlighting key steps while deferring technical details to the Appendix. We start with the proof of \Cref{thm:main_population}.

    \subsection{Sketch Proof of Theorem \ref{thm:main_population}}

    The proof is divided into three cases, depending on the input dimension $m$: when $m>d$, when $2\leq m\leq d$, and when $m=1$. We start with the simplest case: $m>d$.

        \subsubsection{Case 1: $m>d$}
            
            In this case, since there are $m-d$ redundant input variables, we introduce a layer $g_1:\RR^m\to\RR^{d+1}$ that effectively ignores these redundant input variables:
            \[g_1(x_1,...,x_m) = (x_1,...,x_{d+1}).\]
            $g_1$ is a one-layer ReLU neural network with zero bias, and a weight matrix \\
            $W = \begin{bmatrix}
                \mathrm{I}_{(d+1) \times (d+1)} &  \mathrm{0}_{(d+1) \times (m-d-1)}
            \end{bmatrix}.$
            This step simplifies the network's input to $d+1$ dimensions, aligning with the settings of Lemma \ref{lem:Dahal1}.
            We then define the pushforward measure of $\rho$ via $g_1$, defined on $\RR^{d+1}$, by $g_1$ as 
            $\rho' = {g_{1}}_\sharp (\rho)$. By Lemma \ref{lem:Dahal1}, for any $\epsilon>0 $, there exists a neural network $g'\in \mathscr{G}_{NN}(d+1,L',p',\kappa)$ such that
            \[W_1(g'_{\sharp }(\rho'), Q) < \epsilon,\]
where $L' = O\left(\log\left(\frac{1}{\epsilon}\right)\right)$, $p' = O\left(D \epsilon^{-d/\alpha}\right)$, $\kappa = B$.

            Now we let $g\coloneqq g'\circ g_1$, so we have
            \begin{align*}
            W_1(g_{\sharp }(\rho), Q) =W_1((g' \circ g_1)_{\sharp }(\rho), Q)=W_1(g'_\sharp ( {g_1}_{\sharp }(\rho)), Q)=W_1(g'_\sharp ( \rho'), Q)< \epsilon.
            \end{align*}
            Finally, observe that $g\in \mathscr{G}_{NN}(m,L,p,\kappa)$, where
            $L=L'+1= O\left(\log\left(\frac{1}{\epsilon}\right)\right)$, $p=\max\{p',m\} = O\left(D \epsilon^{-d/\alpha}\right)$, $\kappa = \max\{B,1\}$, which finishes the proof of this case.

        \subsubsection{Case 2: $2\leq m\leq d$}
            In this case, where the input dimension lies between 2 and $d$, our strategy comprises several steps. First, we find a $m-1$ dimensional compact Riemannian manifold $\MM'_\epsilon$ approximating $\MM$. Then, we project the original distribution $Q$ onto $\MM'_\epsilon$ to obtain $Q'_\epsilon$, an approximation of $Q$. Next, we smooth $Q'_\epsilon$ into $Q''_\epsilon$ to satisfy Assumption \ref{as:2}. Finally, we show that $Q''_\epsilon$ can be approximated by a deep generative model, as detailed step by step below.

            Firstly, the existence of such a $m-1$ dimensional manifold $\MM'_\epsilon$ is guaranteed by the following Lemma \ref{lem:space_filling} leveraging the idea of a space-filling manifold, with proofs in \Cref{apdx:space_filling}.
            \begin{lemma}\label{lem:space_filling}
                Let $\MM$ be a d dimensional Riemannian manifold isometrically embedded in $\mathbb{R}^D.$ Then, for any $1\leq q \leq d$ and any $\epsilon>0$, there exists a $q$-dimensional manifold isometrically embedded in $\mathbb{R}^D$, denoted by $\MM'_{\epsilon}$ satisfying
            
                \[d(\MM,\MM_\epsilon') = \adjustlimits\sup_{x\in \MM} \inf_{y\in\MM_\epsilon'} \|x-y\| < \epsilon.\]
            \end{lemma}

            Secondly, we define $Q_\epsilon'\coloneqq \pi_{\epsilon_\sharp(Q_\epsilon)}$, where $\pi_\epsilon:\MM\to\MM_\epsilon'$ is the orthogonal projection onto $\MM'_{\epsilon}$. The Wasserstein distance $W_1(Q,Q')$ is then bounded by $W_1(Q,Q_\epsilon')\leq d(\MM,\MM_\epsilon')<\epsilon$.
            
            Thirdly, since $Q_\epsilon'$ may not satisfy Assumption \ref{as:2}, we cannot directly apply Lemma \ref{lem:Dahal1}. To address this, we find another distribution on $\MM'_{\epsilon}$, denoted by $Q''_\epsilon$ that satisfies Assumption \ref{as:2}, with $W_1(Q'_\epsilon,Q''_\epsilon) < \epsilon,$ . The existence of such $Q''_\epsilon$ is supported by the following Lemma \ref{lem:smoothing_measure}, with proofs given in \Cref{apdx:smoothing}.
            
            \begin{lemma}\label{lem:smoothing_measure}
                Let $\MM'$ be a $m-1$ dimensional Riemannian manifold isometrically embedded in $\mathbb{R}^D$, and let $Q'$ be a distribution on $\MM'$. Then, for any $\epsilon>0$, there exists a distribution $Q''$ on $\MM'$ satisfying Assumption \ref{as:2}, with
                \[W_1(Q',Q'') < \epsilon.\]
            \end{lemma}

            Finally, by Lemma \ref{lem:Dahal1}, there exists a network $g\in\mathscr{G}_{NN}(m,L,p,\kappa)$ such that $W_1(g_{\sharp}(\rho), Q''_\epsilon) < \epsilon$, where \[L = O\left(\log\left(\frac{1}{\epsilon}\right)\right),\;\;\; p = O\left(D \epsilon^{-\frac{m}{\alpha(m,\epsilon)}}\right), \;\;\;\kappa = B.\]
           By the triangular inequality, we have
            \begin{align*}
                W_1(g_{\sharp }(\rho),Q) & \leq W_1(g_{\sharp \rho},Q''_\epsilon)+W_1(Q''_\epsilon,Q'_\epsilon)+W_1(Q'_\epsilon,Q)  \leq 3\epsilon.
                \end{align*}

            The remaining step is to show that $\alpha(m,\epsilon)\to 0$ as $\epsilon\to0$ for any $m$, which leads to the super-exponential complexity. However, this proof is technical, so we summarize it in Lemma \ref{lem:alpha} and defer the proofs to \Cref{apdx:alpha}.  
            \begin{lemma}\label{lem:alpha}
                $\alpha(m,\epsilon)\to 0$ as $\epsilon\to0$ for any $2\leq m\leq d$. 
            \end{lemma}

            % \begin{lemma}\label{lem:derivative_explosion}
            %     Let $\MM$ be a d dimensional Riemannian manifold isometrically embedded in $\mathbb{R}^N.$ Let $M_\epsilon'$ be an m dimensional Riemannian manifold with a single chart $g^*$ where $m<d.$ Then, if
            
            %     \[d(\MM,\MM'_{\epsilon}) < \epsilon \to 0\]
            %     We have
            %     \[sup(|Dg^*|) \to \infty\]
            % \end{lemma}
            
            % We can then show that Holder exponent of the local transport $\alpha_i$ approaches 0 as well, and that $\alpha(\epsilon) \to 0$ when $m < d.$ However, it is not known when $m = d.$ The order guarantee may explode, but the true rate may theoretically be lower.

        \subsubsection{Case 3: $m=1$}
            For the case where $m=1$, the approach for $2\leq m\leq d$ is not directly applicable, as Lemma \ref{lem:Dahal1} requires the input dimension to be at least two. Thus, we adopt a different strategy to construct the approximation manifold. Notably, in this case, the input space is a unit interval $[0,1]$, with $\rho=\mathrm{Unif}(0,1)$. 
            
            First, we apply Lemma \ref{lem:space_filling} again, with $q=1$, to obtain an approximation manifold $\MM'_\epsilon$, a one-dimensional space-filling curve. Subsequently, we define $Q'_\epsilon = \pi_{\epsilon\sharp}(Q)$, where $\pi_\epsilon:\MM\to\MM_\epsilon'$ is the orthogonal project. This yields $Q'_\epsilon$ as a distribution on $\MM_\epsilon'$ satisfying $W_1(Q,Q_\epsilon') \leq \epsilon$.
            
            Lemma \ref{lem:reparameterize} guarantees the construction of a smooth parameterization for the curve $\MM'_\epsilon$, given by $\eta_\epsilon: [0,1] \to \MM'_\epsilon$, with $W_1(\eta_{\epsilon\sharp }(\rho),Q_\epsilon') \leq \epsilon.$ The proof is in \Cref{apdx:lem_m=1}.

            \begin{lemma}\label{lem:reparameterize}
                Let $\MM_{\epsilon}'$ be a parametric curve of dimension 1, and let $Q'_\epsilon$ be a distribution on $\MM_{\epsilon}'$, then there exists a reparameterization $\eta_\epsilon:[0,1]\to \MM'_\epsilon$ such that 
                \[W_1(\eta_{ \epsilon\sharp }(\rho),Q_\epsilon') \leq \epsilon.\]
            \end{lemma}
            
            Next, Lemma \ref{lem:Dahal_lem_3} allows us to approximate $\eta$ using a neural network $g$.
            
            %leading to $\|g_\epsilon - \eta\|_\infty \leq \epsilon,$
            %and subsequently 
            %\[W_1(g_{\sharp }(\rho),\eta_{\sharp}(\rho)) \leq \int_{[0,1]} \|g_\epsilon - \eta\| d\rho_\epsilon \leq \epsilon.\]

            %The complexity of $g_\epsilon$ is a direct consequence of \Cref{lem:Dahal_lem_3}, as stated below without a proof. 

            \begin{lemma}[\citet{dahal2022deep} Lemma 3]\label{lem:Dahal_lem_3}
                If $\eta(t)$ is $\alpha$ H\"older continuous and bounded, then for any $\epsilon > 0$ there is a ReLU network $g$ such that
                \[\int_{[0,1]} \|g(t)-\eta(t)\| \mathrm{d}t \leq \epsilon,\]\
                and $g \in \mathscr{G}_{NN}(1,L,p,\kappa)$, where $L=O\left(\log\left(\frac{1}{\epsilon}\right)\right)$, $p=O\left(\epsilon^{-\frac{1}{\alpha}}\right)$, and $\kappa=B$.
            \end{lemma}

            Then we obtain $g$ with $W_1(\eta_{\epsilon\sharp}(\rho),g_{\sharp}(\rho))\leq \epsilon$ and
            \begin{align*}
               W_1(Q,g_{\sharp}(\rho))\leq W_1(Q,Q'_\epsilon)+W_1(Q'_\epsilon,\eta_{\epsilon\sharp }(\rho))+W_1(\eta_{\epsilon\sharp}(\rho),g_{\sharp}(\rho)) < 3\epsilon.
            \end{align*}
        Finally, we node that $\alpha=\alpha(\epsilon)$ depends on $\epsilon$ through $\eta_\epsilon$, and we claim that $\alpha(\epsilon)\xrightarrow[]{\epsilon\to0}0$. This is summarized as Lemma \ref{lem:alpha_m=1} with proofs in \Cref{apdx:alpha_m=1}, which completes the proof of this $m=1$ case. 

     \begin{lemma}\label{lem:alpha_m=1}
                $\alpha(\epsilon)\xrightarrow[]{\epsilon\to0}0$. 
            \end{lemma}

    \subsection{Sketch Proof of \Cref{thm:main_empirical}}
     Let $x_1,...,x_n $ be iid samples from $Q$, and let $Q_n$ denote the associated empirical distribution. By \Cref{thm:main_population}, for any $\epsilon>0$ ,there exists $g\in\mathscr{G}_{NN}(m,L,p,\kappa)$ such that $W_1(Q, g_{\sharp}(\rho))<\epsilon$, where 
            $L=O\left(\log\left(\frac{1}{\epsilon}\right)\right)$, $p=O\left(D\epsilon^\frac{-m}{\alpha(m,\epsilon)}\right)$, and $\kappa = B$.
            
    For the empirical minimizer $\widehat{g}_n$, we can break down the Wasserstein distance $W_1(\widehat{g}_{n\sharp }(\rho),Q)$ as follows:
    \begin{align*}
    W_1(\widehat{g}_{n\sharp }(\rho),Q) &\leq W_1(\widehat{g}_{n\sharp }(\rho),Q_n)+W_1(Q_n,Q)
    \leq W_1(g_{\sharp }(\rho),Q_n)+W_1(Q_n,Q)\\
    &\leq  W_1(g_{\sharp }(\rho),Q)+W_1(Q,Q_n)+W_1(Q_n,Q)
%    &=W_1(g_{\sharp }(\rho),Q)+2W_1(Q,Q_n)\\
    \leq \epsilon + 2W_1(Q,Q_n).
    \end{align*}

    Given Assumptions \ref{as:1} and \ref{as:2}, Lemma 5 of \citet{dahal2022deep} implies that for any $\delta > 0$, a constant $C_\delta$ exists such that
    \[\EE\left[  W_1(Q,Q_n) \right] \leq C_\delta n^{-\frac{1}{d+\delta}},\]
    where $C_\delta$ is a constant independent of $n$. Setting $\epsilon = n^{-\frac{1}{d+\delta} }$, we have
    \[\EE \left [ W_1(\widehat{g}_{\epsilon\sharp }(\rho),Q) \right ] \leq (1+2C_\delta)n^{-\frac{1}{d+\delta}} .\]
    Thus, we have achieved the claimed expected Wasserstein distance with desired complexity of $\widehat{g}_n$: $L=O\left(\log\left(n^{\frac{1}{d+\delta}}\right)\right)$, $O\left(Dn^{\frac{m}{(m+\delta)\alpha(m,n,d,\delta)}}\right)$, and $\kappa = B$. In particular, when $m>d$, $\alpha(m,n,d,\delta)$ remains  constant ($\alpha$), while when $m\leq d$, $\lim_{n\to\infty}\alpha(m,n,d,\delta)=\lim_{\epsilon\to0}\alpha(m,\epsilon)=0$ as per \Cref{thm:main_population}.

\section{Discussion}

    Using the theory of space-filling curves, in this work we establish key relationships between the input dimension, the true dimension of the data manifold, and the approximation error of fitting deep generative networks. Specifically, we have shown that deep generative networks can learn the manifold structure in a generalizable manner regardless of the input dimension, even when the input dimension is as low as one. Furthermore, we quantify the complexity trade-off among network size, input dimension, and approximation error, showing that underestimating the input dimension below the manifold dimension introduces a super-exponential increase in the width of the network. Furthermore, these results did not require any additional assumptions compared to previous work \citep{dahal2022deep}. Toy simulation studies on two- and three-dimensional manifolds provide empirical evidence of GANs learning such space-filling curves from various input dimensions, offering additional explanation to the success of GANs and similar models to learn distributions without explicit need for dimensionality estimation in practice. 
    
    % future direction 
    
    Based on these results, there are multiple potential directions for future work. First, although we have derived statistical guarantees for existence, there is no guarantee on being able to practically fit such models, especially when the input dimension is heavily underestimated. Second, while we have established upper bounds on the complexity necessary to achieve certain approximation errors, identifying lower bounds poses a formidable challenge and remains an open question. The absence of explicit lower bounds in existing literature highlights the complexity and novelty of this problem area. Establishing these bounds would not only deepen our understanding of the behavior of generative networks but also potentially guide the development of more efficient network architectures. Third, exploring the phase transition in model complexity versus approximation error as a potential method for estimating the intrinsic dimension of data manifolds could provide practical insights into dimensionality reduction and the effective training of generative models. Finally, although we have shown that neural networks are capable of learning distributions exhibiting low dimensional manifold structures, there remains situations in which the data do not lie on a manifold of a single intrinsic dimension throughout, as is the case in the union of manifolds hypothesis \citep{brown2022union}. Such extension would further elucidate the ability of these deep generative models to empirically approximate arbitrary datasets in practice. 

%        \section{Statement of Ethics}

%    Our work here is theoretical in nature in providing theoretical support for the empirical success of deep generative networks. Although our work focuses on explaining how these models work, rather than directly contributing to the model itself, the technology itself comes with significant potential societal impact with applications such as being able to generate fake data that closely resembles reality. Our work broadens the scope and elucidates the usability of these generative models, and we would emphasize that such models should be used with care to not introduce potentially harmful biases in data generation. 

%\newpage
%\bibliographystyle{chicago}
%\bibliography{ref}

%\bibliographystyle{agsm}

%\bibliography{ICML-2024/ref}

%\newpage
\appendix
%\section{abc}

%\newpage
%\appendix

% \bigskip
% \begin{center}
% {\large\bf Supplementary Material for ``Deep Generative Models: Complexity, Dimensionality, and Approximation"}
% \setcounter{page}{1}
% \end{center}

\section{Code Availability}\label{apdx:code}%\label{sup:exp}
All code for experiments and generating paper figures can be found in the GitHub repository at \url{https://github.com/hong-niu/dgm24}. All experiments were run on a single machine with an Nvidia RTX 4080 GPU with 16 GB of memory and adapted from the Python Optimal Transport (POT) package \cite{flamary2021pot} for computing the Wasserstein distance.

\section{Additional Proof Details}\label{Appendix:Details}
In this section, we prove five lemmas used in the sketch of the proofs of \Cref{thm:main_population} and \Cref{thm:main_empirical}, namely, Lemma \ref{lem:space_filling}, Lemma \ref{lem:smoothing_measure}, Lemma \ref{lem:alpha}, Lemma \ref{lem:reparameterize}, and Lemma \ref{lem:alpha_m=1}. Note that Lemma \ref{lem:Dahal_lem_3} is directly from Lemma 3 of \citet{dahal2022deep}, so we skip its proof.

\subsection{Proof of Lemma \ref{lem:space_filling}}\label{apdx:space_filling}
We first prove Lemma \ref{lem:space_filling}, which shows that for a compact $d$ dimensional manifold $\MM,$ there exists a smaller dimensional manifold $\MM'_\epsilon$ that approximates the manifold $\MM.$
%\begin{replemma}{lem:space_filling}
%Let $\MM$ be a d dimensional Riemannian manifold isometrically embedded in $\mathbb{R}^D.$ Then, for any $1\leq q \leq d$ and any $\epsilon>0$, there exists a $q$ dimensional manifold isometrically embedded in $\mathbb{R}^D$, denoted $\MM'_{\epsilon}$ satisfying 
 %               \[d(\MM,\MM_\epsilon') = \sup_{x\in \MM} \inf_{y\in\MM_\epsilon'} \|x-y\| < \epsilon.\]
%\end{replemma}

\begin{proof}
Since $\mathbb{R}^D$ is a separable metric space, $\MM\subset\RR^D$ is also separable. Let $E \subset \MM$ be a dense countable subset of $\MM$, enumerated $E = \{e_1,e_2,....\}$. Because $E$ is dense on $\MM$, we have
\[\lim_{n\to\infty}\adjustlimits\sup_{x\in \MM} \inf_{y\in \{e_1,...,e_n\}} \|x-y\|_2^2 = 0.\]
Suppose that we can find a sequence of manifolds $\MM_n$ satisfying for all $n \in \mathbb{N}$,
\[E_n\coloneqq \{e_1,...,e_n\} \subseteq \MM_n.\]
We have 
\[0\leq \adjustlimits\sup_{x\in \MM} \inf_{y\in \MM_n} \|x-y\|_2^2 \leq \adjustlimits\sup_{x\in \MM} \inf_{y\in \{e_1,...,e_n\}} \|x-y\|_2^2.\]
Consequently:
\[\lim_{n\to\infty}\adjustlimits\sup_{x\in \MM} \inf_{y\in \MM_n} \|x-y\|_2^2 = 0.\]
Therefore, for any $\epsilon>0$, there exists a large enough $n$ such that
\[\adjustlimits\sup_{x\in \MM} \inf_{y\in \MM_n} \|x-y\|_2^2 < \epsilon.\]
It suffices to show that for any finite set $E_n$, there exists a Riemannian Manifold $\MM_n$ interpolating $E_n$, that is, $E_n \subseteq \MM_n$. 

We know there exists a $q$ dimensional subspace, represented by the associated projection matrix $P$, such that, for all $i\neq j,~Pe_i \neq P e_j$. We can choose a basis of this subspace to be the first $q$ coordinates through a change of basis. For convenience, we thus claim without loss of generality that the projection onto the first $q$ coordinates $\pi_q(x_1,...,x_n) = (x_1,...,x_q)$ is injective on $E_n$ so that
\[\pi_q(e_i) \neq \pi_q(e_j) \;\text{ when }\; i\neq j.\]
Then there exists infinitely many interpolants, $f$, satisfying $y_i = f(\pi_q(e_i))$. 

Included in our options are smooth interpolants, such as polynomial interpolants, splines, variational minimizing surfaces, RBF interpolations, Backus-Gilbert~\citep{Wendland2004ScatteredDA}, etc. We choose the cubic polyharmonic/thin plate spline due to its simplicity~\citep{duchon1977splines}:

\[f = \min_{f \in W^m_2(\pi_q(M)),y_i = f(\pi_q(e_i))} J_m^d(f),\]
where $J_m^d(f) = \sum_{\alpha_1+...+\alpha_d = m} \frac{m!}{\alpha_1!...\alpha_d!} \int_{-\infty}^\infty ... \int_{-\infty}^\infty \left(  \frac{\partial^m f}{\partial x_1^{\alpha_1}...\partial x_d^{\alpha_d}}  \right) \prod_{j=1}^{d}\mathrm{d}x_j$. 

This interpolant is a smooth function on a compact domain, with a bounded derivative and its image is a smooth manifold. Further, we consider the function:

\[F: \pi_{q}(\MM) \to \mathbb{R}^{n}, (x_1,...,x_q) \mapsto (x_1, ... ,x_q, f_{q+1}(x_1,...,x_q),...,f_{n}(x_1,...,x_q)).\]

Note that the image $F(\pi_{q}(\MM))$ is a Riemannian manifold that interpolates $F(\pi_q(e_i)) = e_i$.
Thus, $E\subseteq F(\pi_q(e_i))$ and 
\[d(F(\pi_q(e_i)),\MM) \leq d_1(E,\MM) \leq \epsilon.\]
Hence, for any $d$ dimensional Riemannian manifold $\MM$ and any $\epsilon>0$, there exists a $q$ dimensional Riemannian manifold $\MM'$, such that 
\[d(\MM,\MM_\epsilon') = \adjustlimits\sup_{x\in \MM} \inf_{y\in\MM_\epsilon'} \|x-y\|< \epsilon.\]    
\end{proof}

\subsection{Proof of Lemma \ref{lem:smoothing_measure}}\label{apdx:smoothing}
We now prove Lemma \ref{lem:smoothing_measure}, which shows that for any distribution on a compact manifold, we may find a sufficiently close distribution that satisfies Assumption \ref{as:2}.

% \begin{lemma}{Smoothing for 5.1.2}
% \begin{replemma}{lem:smoothing_measure}
% Let $\MM'$ be a $m-1$ dimensional Riemannian manifold isometrically embedded in $\mathbb{R}^D$, and let $Q'$ be a distribution on $\MM'$. Then, for any $\epsilon>0$, there exists a distribution $Q''$ on $\MM'$ satisfying Assumption \ref{as:2}, with
%                \[W_1(Q',Q'') < \epsilon.\]
%\end{replemma}

\begin{proof}

The objective of this lemma is to perturb $Q'$ slightly, creating a perturbed distribution $Q''$ that satisfies Assumption \ref{as:2}. The proof involves the following five steps:

\begin{enumerate}
    \item \textbf{Decomposition of the manifold:} Partition the manifold $\MM_{\epsilon}'$ into a finite union of balls, with each as a coordinate chart derived from some Euclidean ball.

    \item \textbf{Local pull back distribution:} Pull back the local distribution $Q'$ through the exponential map.

    \item \textbf{Smooth local distributions:} Smooth each local pull-back distribution to ensure closeness to the original pull-back.

    \item \textbf{Pushforward and reweighting:} Push these local distributions forward to $\MM'$ using their exponential maps. Paste them together to form a new distribution, and re-weight it to account for overlaps.

    \item \textbf{Density adjustment:} Add a small uniform density to ensure a positive lower bound on density and show that this new distribution is sufficiently close to the original distribution and satisfies Assumptions \ref{as:2}.
\end{enumerate}

Since $\MM'$ is compact, the injectivity radius, denoted by $\iota$, is strictly positive~\citep{do1992riemannian}. Fix $0<r<\iota$ and denote $U_{x} = \exp_{x}(B_x(0,r))$, where $\exp_x$ is the exponential map at $x$ and $B_x(0,r)$ is the Euclidean ball in the tangent space $T_{x}\MM'$. Again, due to the compactness, there exists a finite covering $\{\exp_{x_j}(B_{x_j}(0,r)) :  j = 1,2,...,N\}.$ 

Let $Q_{x_j}'$ be the restriction of $Q'$ onto $U_{x_j}$, that is $Q_{x_j}'(S) = \int_{S} \mathds{1} (x \in U_{x_j}) \mathrm{d}Q'(x)$. Since $r<\iota$, the exponential map $\exp_{x_j}$ is invertible on $U_{x_j}$, so the the pullback of $Q'_{x_j}$ through $\exp_{x_j}(\cdot)$, denoted by $P_{x_j}$, a measure on $B_{x_j}(0,r)$, is well-defined.

However, the pull-back measure $P_{x_j}$ may not be smooth enough. To address this, %define a probability measure $W_{x_j}^{-1} P_{x_j}$ in $B_{x_j}B(0,r),$ where $W_{x_j}$ is the total mass of $P_{x_j}$. Let $X \sim W_{x_j}^{-1} P_{x_j}.$ Then, for a sufficiently small $\delta>0,$ let $Y|X=x$ be uniform in the region $B_{x_j}(x,\delta) \cap B_{x_j}(0,r).$ The Radon Nikodym derivative of the smoothed measure is $W_{x_j} f_y(y)$, where $f_y$ is the marginal density of $Y$.
%To do so, 
for $\delta>0$, we consider the function
\[H_\delta(x,y) = C_\delta(x) \mathds{1}\{\|x-y\| < \delta\},\]
where $C_\delta (x)$ is the normalizing constant $\frac{1}{C_\delta(x)}=\int_{B_{x_j}(0,r)} \mathds{1}(\|x-y\| < \delta) \mathrm{d}y $.
%This will ensure that $H_\epsilon(x,y)$ will preserve the total measure of the new function while limiting the Wasserstein distance to below $\epsilon.$ Preserving the total measure ensures that the Wasserstein distance may be computed.

Then denote the locally smoothed measure by $P_{x_j}^\delta$, with density:
\[p_{x_j}^\delta(y) = \int_{B_{x_j}(0,r)} H_{\delta}(x,y) \mathrm{d}P_{x_j}.\]
Since the transport cost between $(P_{x_j},P_{x_j}^\delta)$ is bounded by $\sup_{H_{\delta}(x,y)=1} \|x-y\| \leq \delta, $ we have $W_1(P_{x_j},P_{x_j}^\delta) \leq \delta.$ 

Pushing $P_{x_j}^\delta$ forward through the exponential map onto $\MM'$ yields a locally smoothed distribution on $U_{x_j}$, $\exp_{x_j\sharp}(P_{x_j}^\delta)$. We have:
\begin{align*}W_1(\exp_{x_j\sharp}(P_{x_j}),\exp_{x_j\sharp}(P_{x_j}^\delta)) &= \inf_{\gamma \in \Gamma(\exp_{x_j\sharp}(P_{x_j}),\exp_{x_j\sharp}(P_{x_j}^\delta))} \int_{U_{x_j}\times U_{x_j}} \|x-y\| \mathrm{d}\gamma\\
&= \inf_{\gamma' \in \Gamma(P_{x_j},P_{x_j}^\delta)} \int_{B_{x_j}(0,r)\times B_{x_j}(0,r)} \|\exp_{x_j}(x)-\exp_{x_j}(y)\| \mathrm{d}\gamma'\\
&\leq L_{x_j}\inf_{\gamma \in \Gamma(P_{x_j},P_{x_j}^\delta))} \int_{B_{x_j}(0,r)\times B_{x_j}(0,r)} \|x-y\| \mathrm{d}\gamma ,
\end{align*}
where $\Gamma$ is the coupling and $L_{x_j}$ is the Lipschitz constant of $\exp_{x_j}$ on $\overline{B_{x_j}(0,r)}$~\citep{do1992riemannian}. Note that as $\delta$ approaches 0, $\exp_{x_j}(P_{x_j}^\delta)$ converges in distribution to $Q_{x_j}'$. 

Now we recombine these locally smoothed measures back into a new measure:
\[Q_{\delta,0} = \sum_{j=1}^N \exp_{x_j\sharp}(P_{x_j}^{\delta}).\]
Then let $N(x)$ to be the number of $U_{x_j}$ containing $x$, that is $N(x) = \sum_{j=1}^N \mathds{1}(x\in U_{x_j})$ so that:
\[\lim_{\delta \to 0}Q_{\delta,0}=\lim_{\delta \to 0} \sum_{c =1}^N \exp_{{x_j}_{\sharp}}(P_{x_j}^{\delta}) =\int_{\MM'} N(x) \mathrm{d}Q'(x).\]
To compensate for overlap in the covering, we adjust $Q_{\delta,0}$ to define a new measure
\[Q_{\delta}(A) = \int_{A} \frac{1}{N(x)} \mathrm{d}Q_{\delta,0}.\]
It is clear that as $\delta \to 0,$ $Q_\delta \to_d Q'$. Due to the compactness of $\MM'$, the cost function is bounded and thus \[\lim_{\delta \to 0} W_1(Q',Q_\delta) \to 0.\]

% Now, to satisfy Assumptions 1 and 2, we need to bound the density from above and below as well.

% To observe the upper bound, let $p_{x_j}^\delta(x)$ be the Radon-Nikodym derivative for $P_{x_j}^\delta$. Then $p_{x_j}^\delta(x)$ is upper bounded because

% \[p_{x_j}^\delta(x) = \int_{B(0,r)} H_{\delta}(x,y) dP_{x_j} \leq \int_{B(0,r)} 1 dP_{x_j} \]
% Since our exponential maps have bounded derivative, the final density $q_\delta(x)$ will also be bounded since:
% \[q_{\delta}(x) \leq \sum_{c \in C} \frac{|\exp_{{x_j}_{\sharp}}'|_\infty |p_{x_j}^\delta(x)|_\infty}{n(x)}\]
So far, however, the density of $Q_\delta$, denoted by $q_\delta$ may not be positively lowered bounded, as required by Assumption \ref{as:2}. To fill in the final gap, we conduct the following surgery to $Q_\delta$. 

By the compactness of $\MM'$, the uniform density, denoted by $u(x)$ is well defined on $\MM"$. Let $\theta>0$ be an arbitrary small number, we define $q'' = (1-\theta)q_\delta + \theta u$. Let $Q''$ be the corresponding distribution, we observe that 1):$Q''$ satisfies Assumption \ref{as:2} for any $\delta$ and $\theta$; 2) $Q''$ converges in distribution to $Q'$ when $\delta,\theta\to 0$.  That is, for any $\epsilon$, there exists $\delta>0$ and $\theta>0$ such that $W_1(Q',Q_\delta)<\frac{\epsilon}{2}$ and $W_1(Q'',Q_\delta)<\frac{\epsilon}{2}$, so that
\begin{align*}
W_1(Q'',Q')\leq W_1(Q'',Q_\delta)+W_1(Q_\delta,Q')<\epsilon,
\end{align*}
which closes the proof. 

\end{proof}

\subsection{Proof of Lemma \ref{lem:alpha}} \label{apdx:alpha}
We now prove Lemma \ref{lem:alpha}, which shows that when the input dimension is smaller than the true dimension, the quantity $\alpha(m,\epsilon)$ approaches 0 as $\epsilon$ approaches 0.

%\begin{replemma}{lem:alpha}
%$\alpha(m,\epsilon)\to0$ as $\epsilon\to0$ for any $m$.
%\end{replemma}
\begin{proof}
In existing literature, such as \citet{dahal2022deep}, $\alpha$ is often treated as a constant, since the manifold under consideration is fixed. However, in our framework, both the approximating space-filling manifold $\MM'_\epsilon$ and the approximating distribution $Q'_\epsilon$ change with $\epsilon$. In this situation, it becomes imperative to keep track of the ``constant" $\alpha$ while the manifold $\MM'_\epsilon$ and $Q'_\epsilon$ are evolving. This task is extremly challenging for several reasons. First, much of the existing literature, including \citet{villani2009optimal,dahal2022deep}
, treats $\alpha$ as a constant, without a in-depth discussion. Second, even when we scrutinize their proofs, they typically establish only the existence of $\alpha$, without any explicit construction. Third, the intricacies of $\alpha$ are hidden within the study of Monge-Amp\`ere equations of the potentials of the density functions of $Q'_\epsilon$, a distinct, although related, field of study. 

Our analysis begins with the recent contributions of \citet{dahal2022deep}, who claim that there exists a constant $\alpha$ so that the complexity $p=O(D\epsilon^{-d/\alpha})$. The foundation of their construction of optimal transport relies on Theorem 12.50 from \citet{villani2009optimal}, also known as Caffarelli's regularity theory. Within this theorem, the constant $\alpha$, denoted by $\beta$ by Villani, is introduced ambiguously as ``for some $\beta\in(0,1)$", without further elucidation provided. To address this, we examined a series of papers by Caffarelli~\citep{caffarelli1990interior,caffarelli1990localization,caffarelli1991some,caffarelli1992boundary,caffarelli1992regularity,caffarelli1996boundary}, where we discovered that the constant $\alpha$ depends on the infimum of the density function of $Q_\epsilon'$, denoted by $q_\epsilon'$. In essence, as $\epsilon\to0$, $\inf q_\epsilon'\to0$, leading to the conclusion that $\alpha(m,\epsilon)\to0$. We prove this claim step by step. 

\begin{enumerate}
    \item Prove $\vol(\MM_\epsilon')\xrightarrow[]{\epsilon\to0} \infty$.
    
First we prove that as $\epsilon\to0$, the volume of the space-filling manifold $\MM_\epsilon'$ goes to infinity. Note that this volume is the Riemannian volume of the $m-1$ dimensional manifold $\MM_\epsilon'$, not the volume in the ambient space. By Lemma \ref{lem:space_filling}, the tube around $\MM_\epsilon'$ with radius $2\epsilon$, denoted by $T_{2\epsilon}(\MM_\epsilon')\coloneqq\{x\in\RR^D:d(x,\MM_\epsilon')<2\epsilon\}$, contains $\MM$. Then, by Weyl's Tube Theorem~\citep{weyl1939volume,gray2003tubes}, the volume of the tube is given by the following series expansion:

\vspace{-0.4 in}
%\begin{equation}\label{eqn:Weyl}
%\vol(T_{2\epsilon}(\MM_\epsilon')) = \frac{(\pi2\epsilon^2)^{\frac{D-(m-1)}{2}}}{(\frac{1}{2}(D-(m-1))!}\sum_{l=0}^{[(m-1)/2]}\frac{k_{2l}(\MM_\epsilon')(2\epsilon)^{2l}}{(D-(m-1)+2)(D-(m-1)+4)\cdots(D-(m-1)+2l)},
%\end{equation}

\begin{align*}\label{eqn:Weyl}
&\vol(T_{2\epsilon}(\MM_\epsilon')) = \\
&\frac{(\pi2\epsilon^2)^{\frac{D-(m-1)}{2}}}{(\frac{1}{2}(D-(m-1))!}\sum_{l=0}^{[(m-1)/2]}\frac{k_{2l}(\MM_\epsilon')(2\epsilon)^{2l}}{(D-(m-1)+2)(D-(m-1)+4)\cdots(D-(m-1)+2l)},
\end{align*}

where $k_{2l}(\MM_\epsilon')$ is the integrated mean curvature of $\MM_\epsilon'$ and $k_{0}(\MM_\epsilon')=\vol(\MM_\epsilon')$ as a special case when $l=0$. Focusing on the leading term, we have
\begin{equation}\label{eqn:Weyl_1}
\vol(T_{2\epsilon}(\MM_\epsilon')) = \frac{(\pi2\epsilon^2)^{\frac{D-(m-1)}{2}}}{(\frac{1}{2}(D-(m-1))!}\left(\vol(\MM'_\epsilon)+O(\epsilon^2)\right).
\end{equation}

Similarly, we estimate the volume of the tube around $\MM$, $T_{\epsilon}(M)$ as:
\begin{equation}\label{eqn:Weyl_M}
\vol(T_{\epsilon}(\MM)) = \frac{(\pi\epsilon^2)^{\frac{D-d}{2}}}{(\frac{1}{2}(D-d)!}\left(\vol(\MM)+O(\epsilon^2)\right).
\end{equation}
Now we show that $T_{2\epsilon}(\MM_\epsilon')\supset T_{\epsilon}(\MM)$. For any $x\in T_{\epsilon}(\MM)$, there exists $y\in \MM$ such that $\|x-y\|<\epsilon$. By the construction of $\MM_\epsilon'$, there exists $z\in\MM_\epsilon'$ such that $\|y-z\|<\epsilon$, as a result, 
\begin{align*}
    d(x,\MM_\epsilon') \leq \|x-z\|\leq \|x-y\|+\|y-z\|\leq 2\epsilon,
\end{align*}
which implies $x\in T_{2\epsilon}(\MM_\epsilon')$ and hence $T_{2\epsilon}(\MM_\epsilon')\supset T_{\epsilon}(\MM)$. 

Linking the two volumes in \Cref{eqn:Weyl_1} and \Cref{eqn:Weyl_M}, we conclude that
\begin{align*}
    &~~~~~~~~\frac{(\pi2\epsilon^2)^{\frac{D-(m-1)}{2}}}{(\frac{1}{2}(D-(m-1))!}\left(\vol(\MM'_\epsilon)+O(\epsilon^2)\right) \geq \frac{(\pi\epsilon^2)^{\frac{D-d}{2}}}{(\frac{1}{2}(D-d)!}\left(\vol(\MM)+O(\epsilon^2)\right)\\
    &\Longrightarrow  \vol(\MM_\epsilon')\gtrapprox C\epsilon^{\frac{m-1-d}{2}} \xrightarrow[]{\epsilon\to0}\infty,
\end{align*}
where the last limit is due to $m-1-d<0$. 

\item Prove $\inf q_\epsilon' \xrightarrow[]{\epsilon\to0}0$.

Since $\MM_\epsilon'$ is compact and $q_\epsilon'$ integrates to one, we have
$$1=\int_{\MM_\epsilon'} q_\epsilon' \mathrm{d}\vol_{\MM_\epsilon'}\geq \vol(\MM'_\epsilon)\cdot \inf q_\epsilon'.$$ 
As a result, we conclude that $\inf q_\epsilon' \xrightarrow[]{\epsilon\to0}0$ since $\vol(\MM_\epsilon')\xrightarrow[]{\epsilon\to0} \infty$.

\item Prove $\alpha(m,\epsilon)\xrightarrow[]{\epsilon\to0}0$.

As stated in Theorem 12.50 of \citet{villani2009optimal}, the optimal transport mapping, denoted by $F_\epsilon$, is represented by its potential $F_\epsilon\in C^{1,\alpha(m,\epsilon)}$ such that $F_\epsilon=\nabla \psi_\epsilon\in C^{\alpha(m,\epsilon)}$. Moreover, we know that $q_\epsilon'(\nabla\psi_\epsilon)\det D_{ij}\psi_\epsilon=q$ from Equation (1) of \citet{caffarelli1992regularity}, where $D_{ij}$ represents the Hessian matrix. The existence of such $\alpha(m,\epsilon)$ was first given by \citet{caffarelli1991some}, which replies on the key assumption that there exists constants $\lambda_1,\lambda_2$ such that
\begin{equation}\label{eqn:Caffarelli_assumption}
0<\lambda_{1,\epsilon}\leq \det D_{ij}\psi_\epsilon\leq \lambda_{2,\epsilon}<\infty.
\end{equation}
Then, the key quantity we are interested, is defined as $\alpha(m,\epsilon) = -\log_2\delta(\epsilon)$, where $\delta(\epsilon)<1$ is the constant such that $h_{\epsilon,1/2}(x-x_0)<\delta(\epsilon) <h_{\epsilon,1}(x-x_0)$, where $h_{\epsilon,\beta}$ is the cone generated by $x_0$ and the level surface $\psi_\epsilon=\beta$ (see Lemma 2 of \citet{caffarelli1991some} for more details). The existence of such $\delta(\epsilon)$ is given by Lemma 2 of \citet{caffarelli1991some}. Since $\inf q_\epsilon'\xrightarrow[]{\epsilon\to0}$, we know that $\lambda_{2,\epsilon}=\sup\det D_{ij}\psi_\epsilon \to \infty$, which violates the key assumption in \citet{caffarelli1991some}, leading to the fact that $\psi_\epsilon$ has infinite derivative and hence infinite local changes and inseparable level sets, so that $\delta(\epsilon)\xrightarrow[]{\epsilon\to0}1$. As a result $\alpha(m,\epsilon)=-\log_2\delta(\epsilon)\xrightarrow[]{\epsilon\to0}0$, as desired. 
\end{enumerate}

\end{proof}

\subsection{Proof of Lemma \ref{lem:reparameterize}}\label{apdx:lem_m=1}
We now prove Lemma \ref{lem:reparameterize}, which states that when an approximating curve for a manifold may be reparameterized by a differentiable function such that the pushforwards of the uniform distribution is close to the true distribution.

%\begin{replemma}{lem:reparameterize}
%                 Let $\MM_{\epsilon}'$ be a parametric curve of dimension 1, and let $Q'_\epsilon$ be a distribution on $\MM_{\epsilon}'$, then there exists a reparameterization $\eta_\epsilon:[0,1]\to\MM_\epsilon'$ such that 
%                \[W_1(\eta_{\epsilon \sharp }(\rho),Q_\epsilon') \leq \epsilon.\]
%\end{replemma}

\begin{proof}
We first introduce the arc-length parameterization as $\xi(t):[0,L_\epsilon] \to \MM_{\epsilon}'$, which makes $\xi$ invertible with a constant speed $\|\xi'(t)\| \equiv 1$. For simplicity, we assume the total length $L_\epsilon=1$ in this proof. Let $P_\xi$ be the pull back of $Q_{\epsilon}'$ onto $[0,1]$ through $\xi$, that is, $\xi_\sharp(P_\xi) = Q_\epsilon'$. Then for $\delta>0$, define
\[H_\delta(x,y) = C_\delta(x) \mathds{1}\{\|x-y\| < \delta\},\]
where $C_\delta(x)$ is the normalizing constant:
\[\frac{1}{C_\delta(x)} = \int_{[0,1]} \mathds{1}\{\|x-y\| < \delta\} \mathrm{d}y.\]

Then we define the following density function
\[f_{\delta}(x) \coloneqq  \int_{[0,1]} H_{\delta}(x,y) \mathrm{d}P_\xi \leq \int_{[0,1]} \sup_{x\in [0,1]}\|C_\delta(x)\| \mathrm{d}P_\xi = \sup_{x\in [0,1]}\|C_\delta(x)\|= \frac{1}{\delta}.\]

Denote the corresponding distribution as $F_\delta$. Let $X \sim P_\xi$ and define $Y\mid X=x$ to be uniform on the interval $[x-\delta,x+\delta] \cap [0,1]$, so that $(X,Y) $ is a coupling in $\Gamma(P_\xi, F_\delta)$. The cost of such a transport is bounded by

\[\EE_{(X,Y)}\|X-Y\| = \EE_X\left[\EE_Y\left[\|X-Y\| \mid X\right]\right]\leq \EE_X\left[\EE_Y\left[\delta \mid X\right]\right] = \delta.\]

By the definition of the Wasserstein distance, $W_1(F_\delta, P_\xi) <\delta$. Subsequently, we can bound
\begin{align*}
W_1(\xi_{\sharp}(F_\delta), Q_\epsilon') &= \inf_{\gamma \in \Gamma(\xi_{\sharp}(F_\delta), Q_\epsilon')} \int_{\MM_{\epsilon}' \times \MM_{\epsilon}'} \|x-y\| \mathrm{d}\gamma = \inf_{\gamma \in \Gamma(\xi_{\sharp}(F_\delta), \xi_{\sharp}(P_\xi))} \int_{\MM_{\epsilon}' \times \MM_{\epsilon}'} \|x-y\| \mathrm{d}\gamma\\
&= \inf_{\gamma \in \Gamma(F_\delta, P_\xi)} \int_{[0,1] \times [0,1]} \|\xi(x)-\xi(y)\| \mathrm{d}\gamma
\leq \inf_{\gamma \in \Gamma(F_\delta, P_\xi)} \int_{[0,1] \times [0,1]} \|x-y\| \mathrm{d}\gamma \\
&= W_1(F_\delta, P_\xi)\leq \delta.
\end{align*}

If the density $f_\delta(x)$ is lower bounded away from 0, then the Cumulative Distribution Function (CDF) of $f_\delta$, denoted by $C_\delta(x),$ is invertible. Then we use the inverse transform to approximate $Q_{\epsilon}'$: 
\[W_1((\xi \circ C_\delta^{-1})_{\sharp}(\rho), Q_\epsilon') \leq \delta.\] 
Thus, $\eta_\epsilon(t) \coloneqq \xi(F_\delta^{-1}(t))$ is our desired parametrization. Here, we highlight its dependence on $\epsilon$ through the domain $\MM_\epsilon'$, which is crucial for Lemma \ref{lem:alpha_m=1}. 

If $f_\delta$ is not lower bounded away from 0, we can adopt the same technique in step 5 in the proof of Lemma \ref{lem:smoothing_measure}, where we perturb it by a small uniform distribution with density $u$: $f_\delta^{'}(x) = (1-\theta)f_\delta(x)+u$, where $\theta>0$. Denoting the CDF of $f_\delta'$ as $C_\delta'$, we define $\eta_\epsilon(t) \coloneqq \xi({{C_\delta^{'}}}^{-1}(t))$. 
As a result, there exists $\delta,\theta>0$ such that
\[W_1(\eta_{ \sharp}(\rho), Q_\epsilon')=W_1((\xi \circ C_\delta^{-1})_{\sharp}(\rho), Q_\epsilon') < \epsilon.\]

% Finally, note that $\eta$ has bounded derivative and is thus Holder continuous

% \[\frac{\mathrm{d} \eta(t)}{\mathrm{d}t} \leq \sup_{s\in[0,1]}\left\|\frac{\mathrm{d} \eta(s)}{\mathrm{d}s}\right\| \sup_{t\in[0,1]}\left\|\frac{\mathrm{d} {F_\delta^{'}}^{-1}(t)}{\mathrm{d}t}\right\| = \frac{1}{\inf_{t\in[0,1]} f_\delta'(t)}\]

\end{proof}

\subsection{Proof of Lemma \ref{lem:alpha_m=1}}\label{apdx:alpha_m=1}
We now explain the proof of Lemma \ref{lem:alpha_m=1}, which shows that $\lim_{\epsilon\to 0} \alpha(m,\epsilon) = 0$ when $m=1.$
 
%\begin{replemma}{lem:alpha_m=1}
%    $\alpha(\epsilon)\xrightarrow[]{\epsilon\to0}0$. 
%\end{replemma}

\begin{proof}
The proof follows the same structure as the proof of Lemma \ref{lem:alpha}, with exactly the same 3 steps. In fact, the arguments presented in the proof of Lemma \ref{lem:alpha} are applicable for any dimension $m$. 
\end{proof}

\subsection{Proof of Corollary \ref{corollary:wasserstein-p}}\label{apdx:wasserstein_p}

\changedreviewertwo{In \citet{dahal2022deep}, the authors first prove the existence of an oracle map $g$ that perfectly transports the input distribution to the target distribution, that is $Q = g_{\sharp}(\rho).$ They then approximate this function $g$ via a neural network $f$ such that $\|f - g\|_{L_1(\rho)}\leq \epsilon$, with explicit complexity bound on $f$. Then, in Lemma 2 of their paper, they show that one can further bound the Wasserstein-1 distance by $W_1(f_{\sharp}(\rho), g_{\sharp}(\rho)) \leq C\|f - g\|_{L_1(\rho)}\leq C\epsilon$. }

\changedreviewertwo{To generalize the setting to Wasserstein-$p$ distances, we need two components. First we extend Lemma 2 of \citet{dahal2022deep} to bound Wasserstein-$p$ distances by $\|f-g\|_{L_p(\rho)}$. Then we find a suitable neural network $f$ to control $\|f-g\|_{L_{p}(\rho)}$. We start with the Wasserstein-$p$ distance:}

            \changedreviewertwo{\[W_p(f_\sharp(\rho),g_\sharp(\rho)) = \inf_{\pi \sim \Gamma(f_\sharp(\rho),g_\sharp(\rho))} \EE_{(x,y)\sim \pi}(\|x-y\|_2^p)^{1/p}.\]}

            \changedreviewertwo{Because the Wasserstein-$p$ distance is the infimum over all couplings, we may find an upper bound by using any choice of coupling. The most obvious choice of coupling is $(f(z),g(z)) \sim (f_\sharp(\rho),g_\sharp(\rho))$ where $z\sim \rho,$ which yields the following upper bound:}

            \changedreviewertwo{\begin{align*}
            W_p(f_\sharp(\rho), g_\sharp(\rho))^p &= \inf_{\pi \in \Gamma(f_\sharp(\rho), g_\sharp(\rho))} \mathbb{E}_{(x, y) \sim \pi} \|x - y\|_2^p  \\
            &\leq \mathbb{E}_{z \sim \rho} \|f(z) - g(z)\|_2^p  \\
            &\leq \int \|f(x) - g(x)\|_2^p \, d\rho \\
 %           &\leq \int C^p \|f(x) - g(x)\|_1^p \, d\rho \\
            &= \|f-g\|_{L_p(\rho)}^p.
            \end{align*}}

\changedreviewertwo{According to \citet{dahal2022deep}, the oracle $g$ may be constructed so as to be $\alpha$ H\"older continuous, and thus continuous. The Universal Approximation Theorem of neural networks \cite{hornikapprox} states that if $g$ is continuous, we may construct a neural network $f$ such that $\|f - g\|_{\infty} \leq \epsilon.$ Note that $\|f-g\|_{L_p(\rho)} \leq \|f-g\|_{\infty},$ so we have}
\changedreviewertwo{\[W_p(f_\sharp(\rho), g_\sharp(\rho))\leq \|f-g\|_{L_p(\rho)}\leq \|f-g\|_\infty\leq \epsilon,\]}
\changedreviewertwo{since $\rho$ is a probability measure.} 

\changedreviewertwo{For the empirical result, we follow the same line of argument as in the Wasserstein-1 case. Define $Q_n$ to be the empirical distribution and $\widehat{g}_{n\sharp }$ to be the empirical risk minimizer. Because $Q$ is bounded, we know it has a finite $p$-th moment, so we have}
\changedreviewertwo{\begin{align*}
    W_p(\widehat{g}_{n\sharp }(\rho),Q) &\leq W_p(\widehat{g}_{n\sharp }(\rho),Q_n)+W_p(Q_n,Q)
    \leq W_p(g_{\sharp }(\rho),Q_n)+W_p(Q_n,Q)\\
    &\leq  W_p(g_{\sharp }(\rho),Q)+W_p(Q,Q_n)+W_p(Q_n,Q)
%    &=W_1(g_{\sharp }(\rho),Q)+2W_1(Q,Q_n)\\
    \leq \epsilon + 2W_p(Q,Q_n).
\end{align*}}

\changedreviewertwo{From \cite{weed2019sharp}, we know that if the Minkowski dimension of $Q$ is $\gamma$, then for all $\delta>0,$ there exists $C_\delta>0$ such that $\mathbb{E}(W_p(Q,Q_n)) \leq C_\delta n^{\frac{1}{\gamma+\delta}}.$ By assumption, $Q$ is supported on a compact $d$ dimensional Riemannian manifold and consequently its Minkowski dimension is $d$. Thus, by choosing $\epsilon = n^{-\frac{1}{d+\delta}}$ as we did before, we obtain $\EE \left [ W_p(\widehat{g}_{n\sharp }(\rho),Q) \right ] \leq (1+2C_\delta)n^{-\frac{1}{d+\delta}}$.}

\changedreviewertwo{While we have proved the existence of an approximating deep neural network for the Wasserstein-$p$ distance, providing specific bounds on the network's complexity analogous to those established for the Wasserstein-$1$ distance in \Cref{thm:main_population} and \Cref{thm:main_empirical} remains unresolved. In the case of $W_1$, the complexity bounds are intricately tied to the construction detailed in Lemma 2 of \citet{dahal2022deep}. Extending these bounds to $W_p$ distances would require developing a similar foundational lemma tailored to $W_p$ settings, a task that involves substantial theoretical innovation and is not immediately straightforward. Another potential approach involves leveraging results from \citet{ohn2019smooth}, which offer complexity bounds in the $L_p$ norm; however, this approach introduces a new challenge as the bound on network weights, $\kappa$, becomes dependent on $\epsilon$. Removing the dependency on $\epsilon$ in the bound for network weights is not straightforward, requiring further investigation. Whether we develop constructions similar to Lemma 2 of \citet{dahal2022deep} for $W_p$ distances, or apply theories like those in \citet{ohn2019smooth}, the challenges are nontrivial. We consider this an important direction for future research.}

%\clearpage
\section{Additional Experiments}\label{apdx:exp}

Below we provide additional simulations elucidating the behavior of the neural networks within the same toy scenarios based on the simulations of the main paper to further support our theoretical findings. In Section \ref{apdx:sim1}, we present an additional experiment regarding neural network complexity and approximation error in the context of Simulation 1. \changedreviewerone{In Section \ref{apdx:sim_m=d+1}, we demonstrate consistent findings with the known theory extending each of the 3 simulations of the main paper to the input $m=d+1$ case as proven previously in \citep{dahal2022deep}. In each case we find that a smaller or equivalently sized neural network can learn comparatively close approximations to the $m=d$ cases presented in the main paper.} \changedreviewertwo{Finally, in Section \ref{apdx:sim_normal-distr}, we show that using a normally distributed input, rather than uniform, leads again to equally strong approximation performance against a normally distributed target output.} Again, complete implementation notebooks generating each of the paper figures can be found at: \url{https://github.com/hong-niu/dgm24}.

\subsection{Extensions to Simulation 1 (Unit Square), $m=1, d=2$}\label{apdx:sim1}
    
    In this section, we provide an additional study of complexity size and approximation accuracy when varying the size of the network. We repeat the same setup as Simulation 1 of the main paper as a representative example, where the input dimension $m=1$ and the target data dimension $d=2$, with the following changes. We present the results of three different sized networks a) 2 hidden layers, 10 nodes each, b) 3 hidden layers, 100 nodes each, and c) 5 hidden layers, 200 nodes each, and also significantly increase the training epochs to a maximum of 20,000 iterations for each experiment. In each case, the input distribution is uniform in 1-dimension, and the data distribution is uniform in 2-dimensions. We observe that the approximation improves with increasing network complexity, and that aligned with our theory, a model as small as network a) with two hidden layers is sufficient for fitting the data manifold when the input dimension is equal to the data, but such network is insufficient for approximating the manifold to the same degree when the input dimension is underestimated. 
    
    \begin{figure}[ht!]
            \centering

            \begin{tabular}{c c c}
                \includegraphics[width=.30\textwidth]{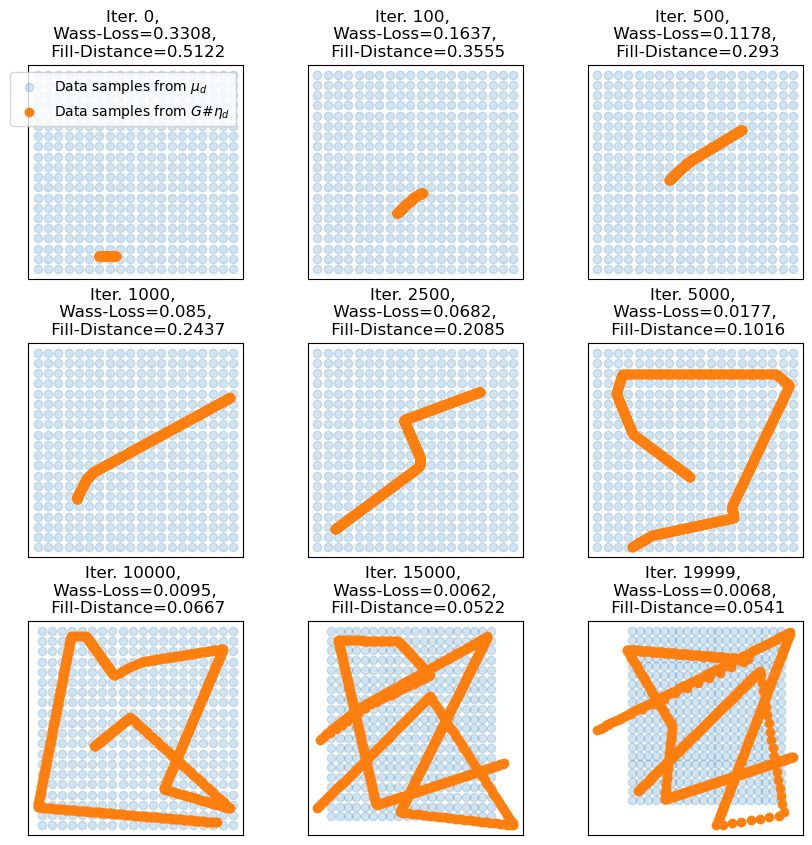} & 
                \includegraphics[width=.30\textwidth]{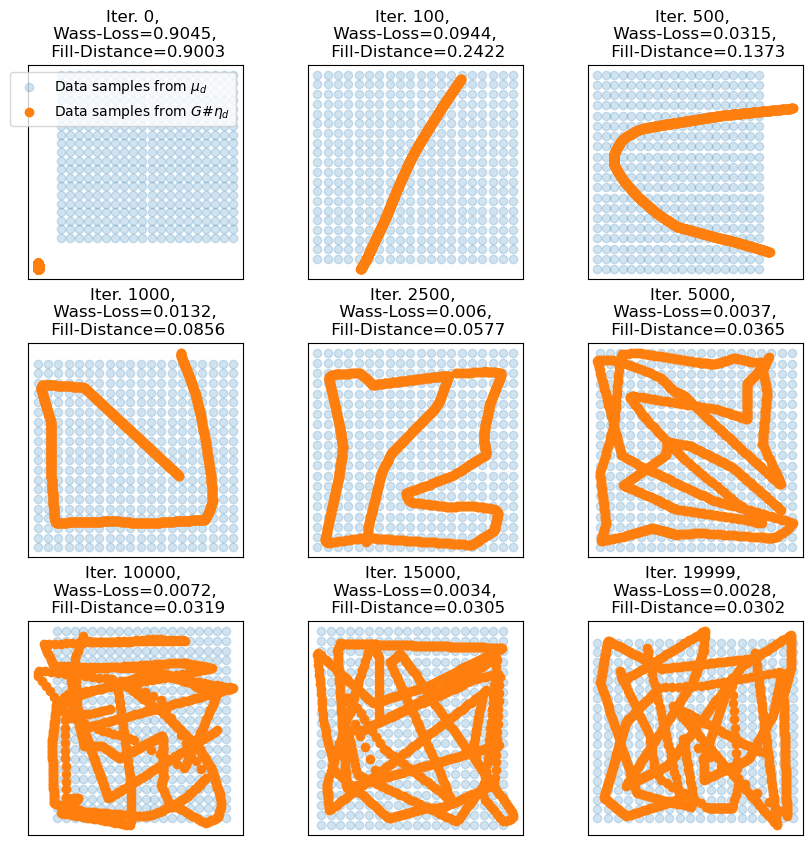} & 
                \includegraphics[width=.30\textwidth]{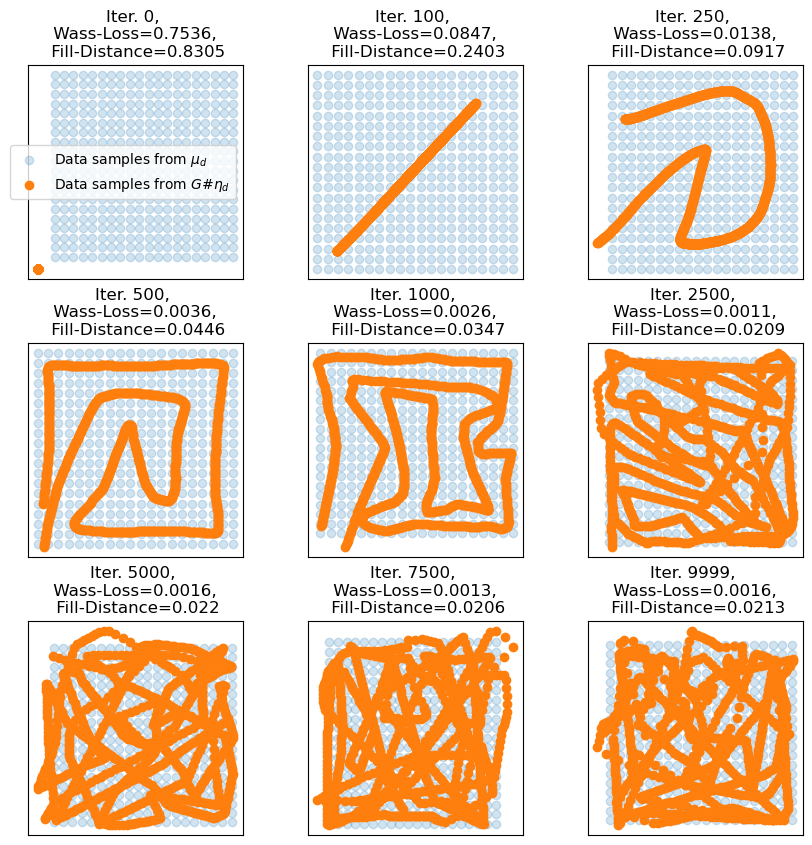} \\
                a) 2 layers, 10 nodes each & b) 3 layers, 100 nodes each & c) 5 layers, 200 nodes each
            \end{tabular}
            \caption{Training results of 3 different network sizes on mapping the 1-dimensional uniform input to a higher 2-dimensional uniform target data manifold. Here the orange curve is generated by the neural network and is ``filling in'' the data manifold (blue).  Networks a) and b) were trained to 20,000 iterations, while network c) only required 10,000 iterations to converge. We observe that as the network complexity increases, the model is able to better fit the target distribution with both lower Wasserstein loss and empirical fill distance against the true data. Note that the smallest network a) is sufficient when the input and target data dimension are equal, as was in Section \ref{sec:sim1}, but is unable to achieve the same degree of approximation after 20,000 training iterations now that the data dimension has been underestimated, aligning with expectations from our presented theory.}
            \label{fig:apdx:sim1-1d2d-fills}
            \vspace{-10pt}
            \end{figure}
            
    Again consistent with our theory, we find that a much larger network c) with 5 hidden layers of 200 nodes each is required to achieve a closer Wasserstein loss and fill distance to the results of Section \ref{sec:sim1} now that the data dimension has been strictly underestimated, with a final Wasserstein loss of 0.0016 and fill distance of 0.0213. Furthermore, network c) improves on the medium sized network in network b) which achieves a final loss of 0.0028 and fill distance of 0.0302. We also note that the largest network c) also consistently achieves lower losses throughout training than the middle sized network b), despite being trained for 10,000 fewer iterations.

        \begin{figure}[ht!]
            \centering

            \begin{tabular}{c c c}
                \includegraphics[width=.30\textwidth]{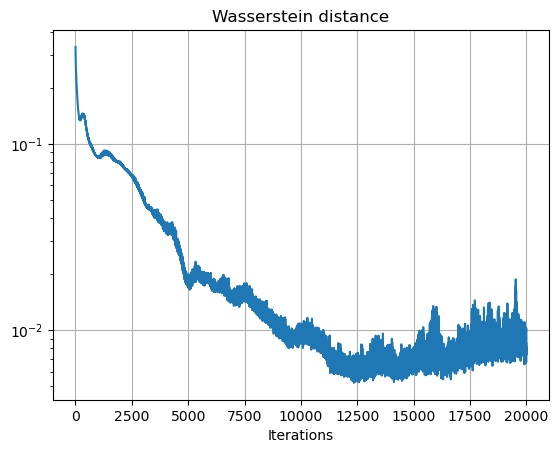} & 
                \includegraphics[width=.30\textwidth]{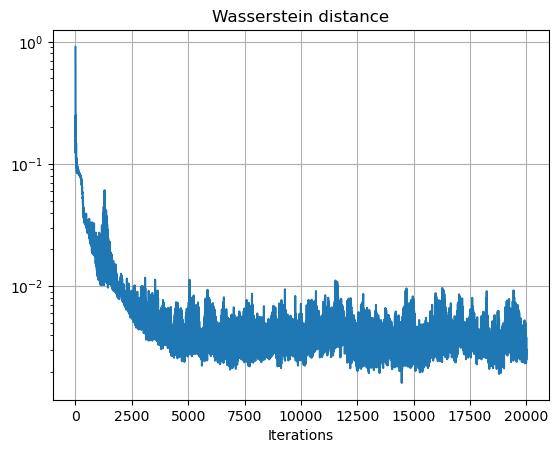} & 
                \includegraphics[width=.30\textwidth]{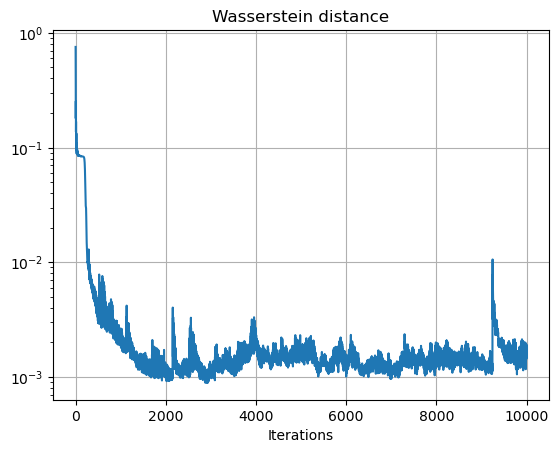} \\
                \includegraphics[width=.30\textwidth]{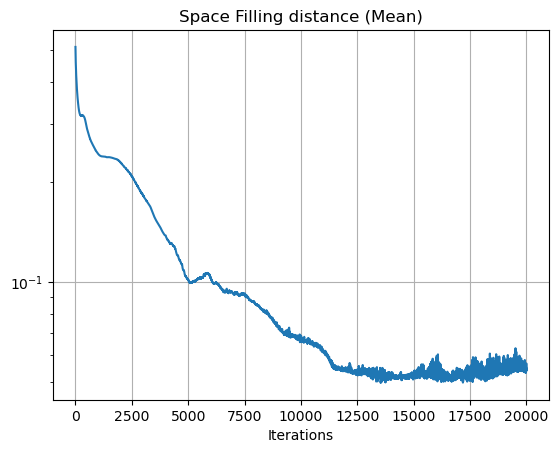} & 
                \includegraphics[width=.30\textwidth]{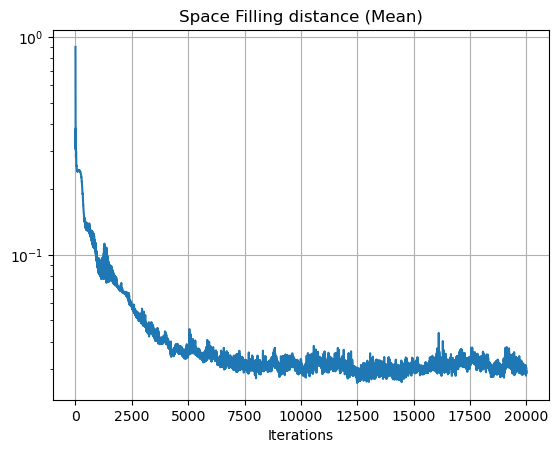} & 
                \includegraphics[width=.30\textwidth]{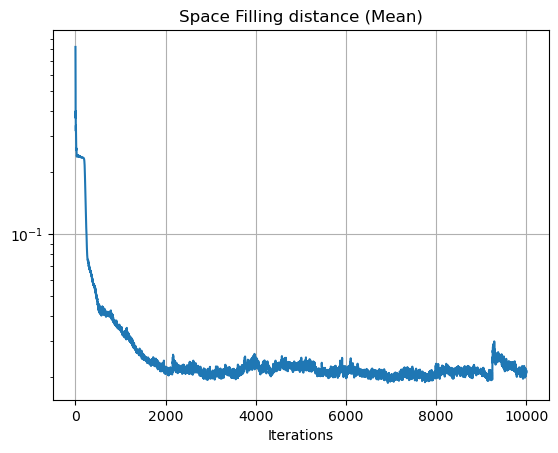} \\
                a) 2 layers, 10 nodes each & b) 3 layers, 100 nodes each & c) 5 layers, 200 nodes each

            \end{tabular}
            \caption{[Top Row] Wasserstein loss per training iteration for each of the three networks presented in Section \ref{apdx:sim1}. [Bottom Row] Fill distance between generated curve and data per iteration for each of the three networks presented in Section \ref{apdx:sim1}.}
            \label{fig:apdx:sim1-1d2d-fills_Wasserstein_Losses}
            % \vspace{-10pt}
            \end{figure}

\clearpage
\subsection{Extensions to the Case for $m=d+1$}\label{apdx:sim_m=d+1}
    \changedreviewerone{Next we show that in each of the cases of Simulation 1, 2 and 3 from the main paper, when the input dimension $m$ is 1 greater than the dimension of the target manifold $d$, a neural network can be trained to approximate the manifold as well as the $m=d$ case as is consistent with previous theory in \citep{dahal2022deep}. We note that we were able to find such a neural network using fewer or equivalently many neurons than that of the $m=d$ case in each of the three cases.} \\

\begin{figure}[h!]
            \centering
            \includegraphics[width=.7\textwidth]{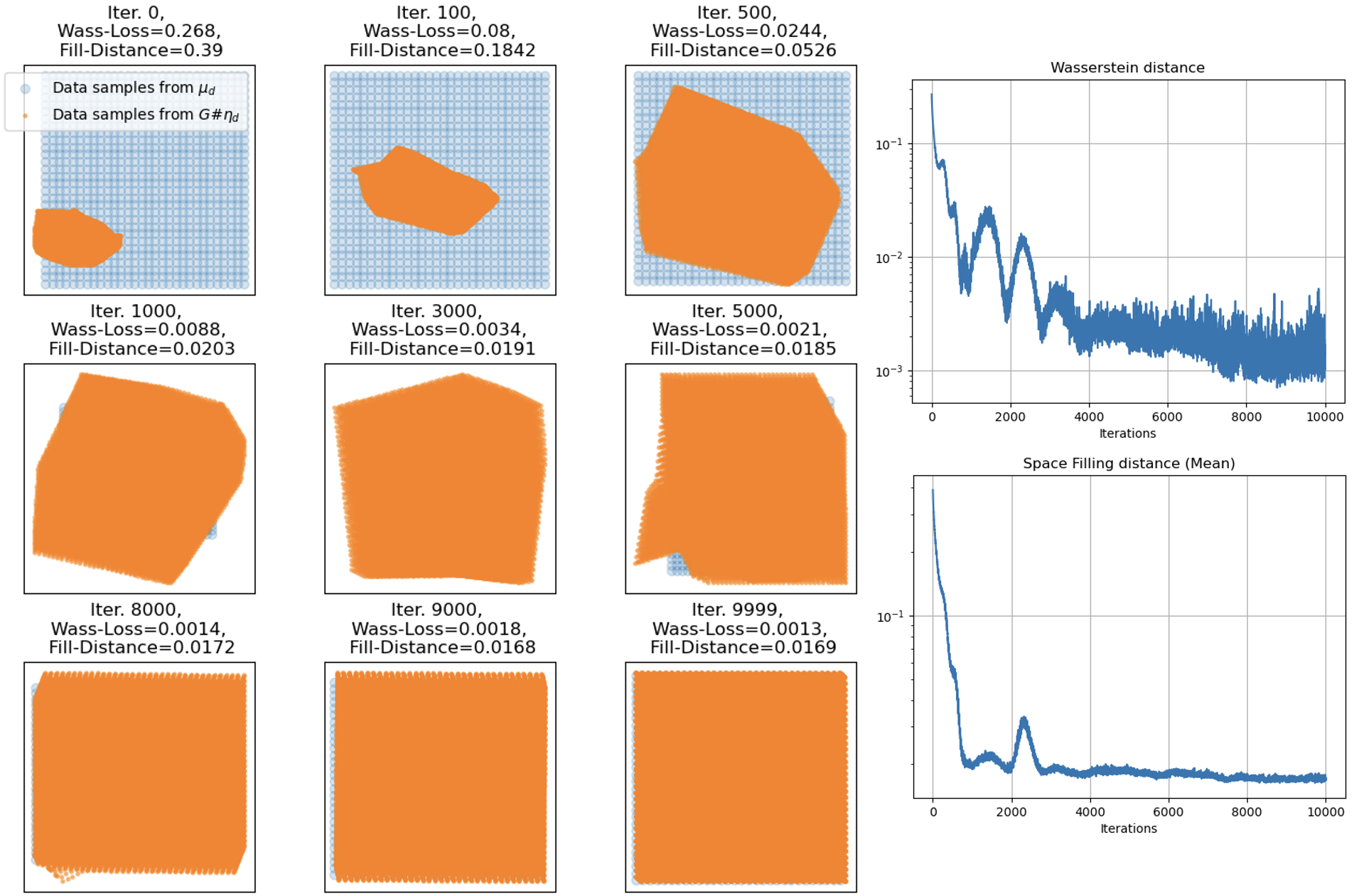}
            \caption{Training results of a 1 hidden layer, 10 node, fully connected network for 10,000 iterations. Training trajectory (left), Wasserstein loss (top right), and fill distance (bottom right) of mapping a uniform distribution on the cube $[0,1]^3$ to a uniform distribution on the unit square $[0,1]^2$. It can be seen that as the loss decreases, the fitted curve fills more of the square as expected, and as well as the 2-D to 2-D analogue in Simulation 1.}
            \label{fig:apdx:sim1-R3R2-square}
            % \vspace{-10pt}
        \end{figure}

    \changedreviewerone{In Figure \ref{fig:apdx:sim1-R3R2-square}, we show that a smaller neural network than the 2-D to 2-D analogue of Simulation 1 is able to learn a mapping between a 3-D uniform input distribution to a smaller 2-D target distribution. We note that in this case, the number of training samples used to train the network (900) is identical to the 2-D to 2-D case in Simulation 1 of the main paper (900), however the number of visualization points for the 3-D uniform input has been increased to maintain the density of the input grid as the 2-D case.}

\begin{figure}[h!]
            \centering
            \includegraphics[width=.7\textwidth]{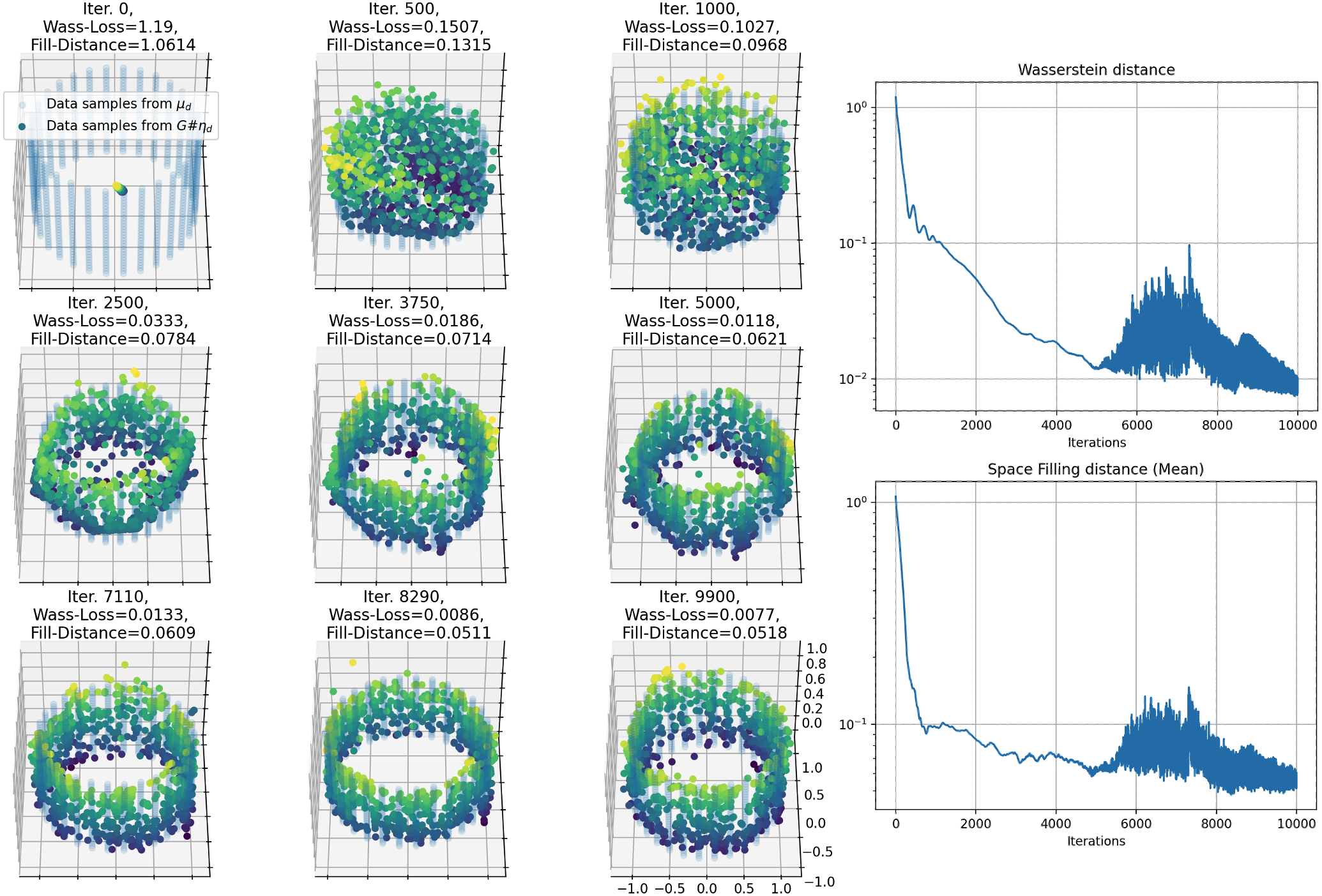}
             \caption{Training results of a 3 hidden layer, 20 node each, fully connected network for 10,000 iterations. Training trajectory (left), Wasserstein loss (top right), and fill distance (bottom right) of mapping a uniform distribution on the unit cube $[0,1]^3$ to a 2-D uniform distribution on a cylinder. It can be seen that as the loss decreases, the fitted curve fills more of the cylindrical surface as expected, and as well as the 2-D to 2-D analogue in Simulation 2.}
            \label{fig:apdx:sim2-R3R2-cylinder}
            % \vspace{-10pt}
        \end{figure}

\begin{figure}[h!]
            \centering
            \includegraphics[width=.7\textwidth]{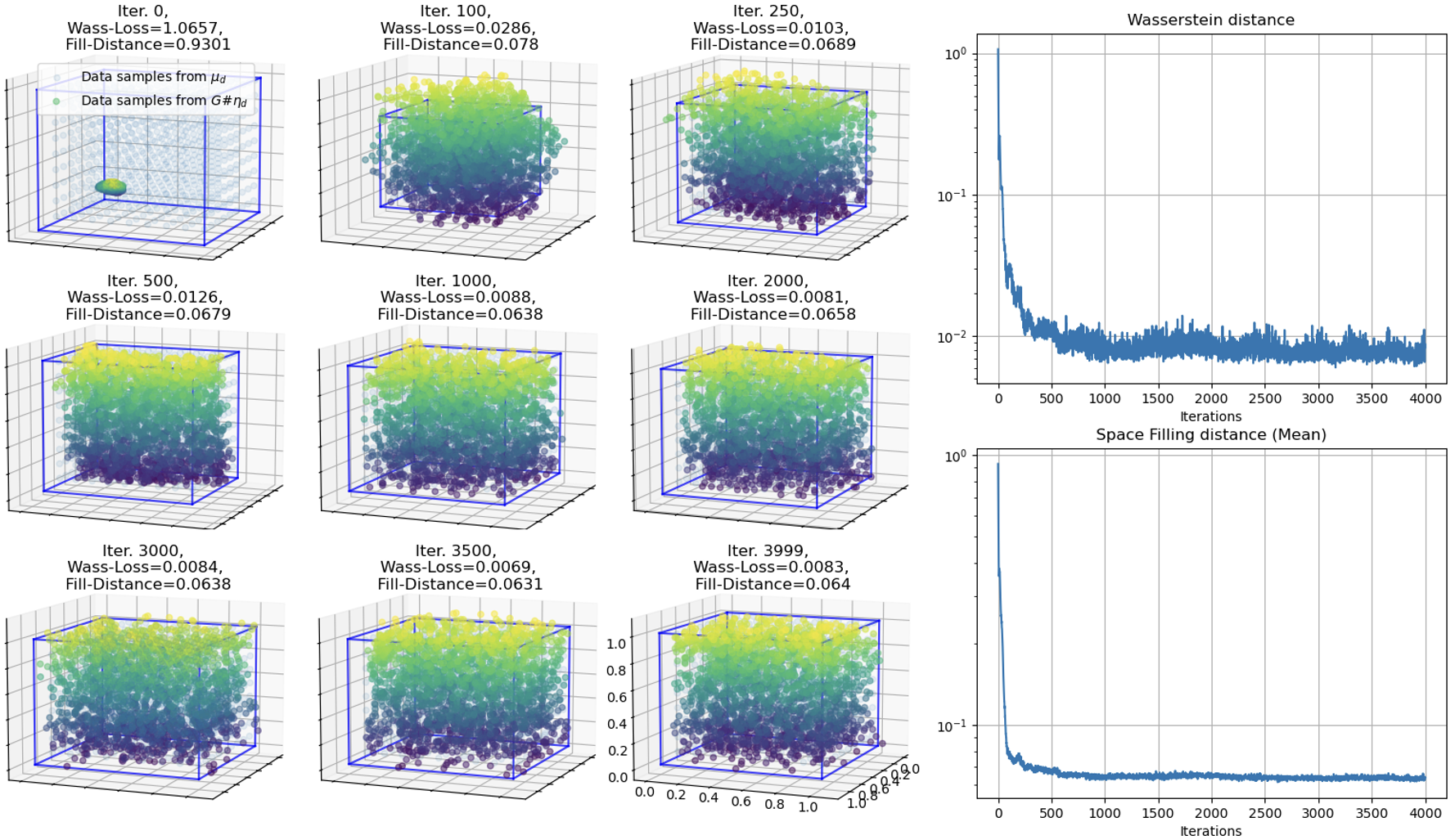}
            \caption{Training results of a 2 hidden layer, 128 node each, fully connected network for 10,000 iterations. Training trajectory (left), Wasserstein loss (top right), and fill distance (bottom right) of mapping a uniform distribution on  $[0,1]^4$ to a uniform distribution on the unit cube $[0,1]^3$. It can be seen that as the loss decreases, the fitted curve fills more of the cube as expected, and as well as the 3-D to 3-D analogue in Simulation 3.}
            \label{fig:apdx:sim3-R4R3-cube}
            % \vspace{-10pt}
        \end{figure}

    \changedreviewerone{Next, we repeat for $m=d+1$ input dimension analogues of Simulation 2 in Figure \ref{fig:apdx:sim2-R3R2-cylinder}, and for Simulation 3 in Figure \ref{fig:apdx:sim3-R4R3-cube}. Again, the number of training samples used in both of these simulations is equal to the respective $m=d$ cases of the main paper, while the neural networks used in the $m=d+1$ cases actually use fewer total neurons to learn similarly accurate approximations.   }

     %fix cylinder 

\clearpage
\subsection{Extensions to Non-Uniform Distributions}\label{apdx:sim_normal-distr}
    \changedreviewertwo{Here, we briefly consider the case when the distributions are no longer uniform. We present two simulations where the target manifold is a 2-D normally distributed sample, and show that keeping the neural network architecture identical and only varying using either a 1-D uniform or 1-D normal distribution as the input results in equally accurate approximations of the target manifold in terms of Wasserstein distance and fill distance as is consistent with our theoretical assumptions.}

      \begin{figure}[h!]
            \centering
            \includegraphics[width=.55\textwidth]{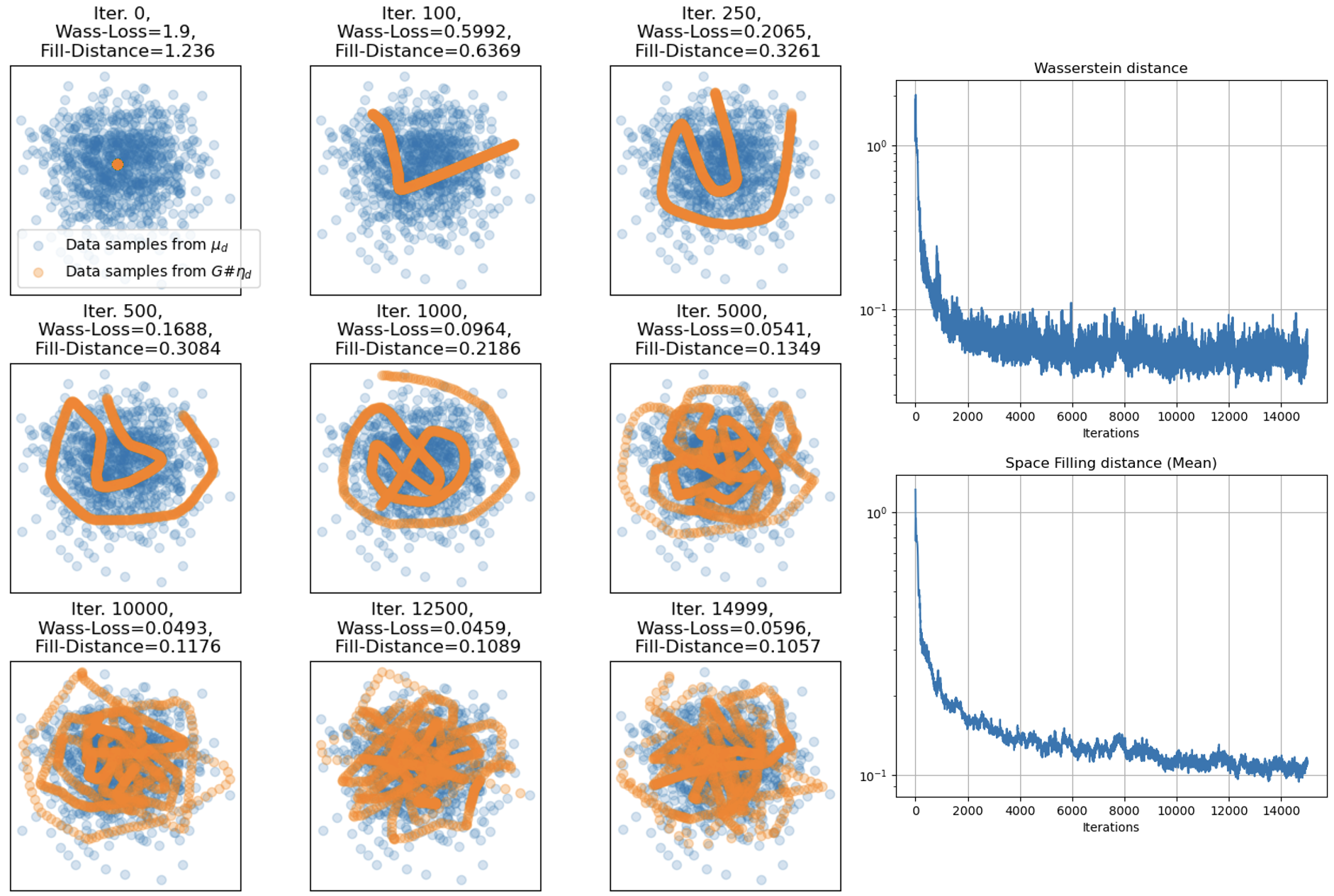}
            \caption{Training results of a 5 hidden layer, 200 node each, fully connected network for 10,000 iterations. Training trajectory (left), Wasserstein loss (top right), and fill distance (bottom right) of mapping a uniform distribution on the interval $[0,1]$ to a 2-D normally distributed target sample.}
            \label{fig:apdx:sim-1-trajectory-unif-normal}
             \vspace{-10pt}
        \end{figure}

        \begin{figure}[h!]
            \centering
            \includegraphics[width=.55\textwidth]{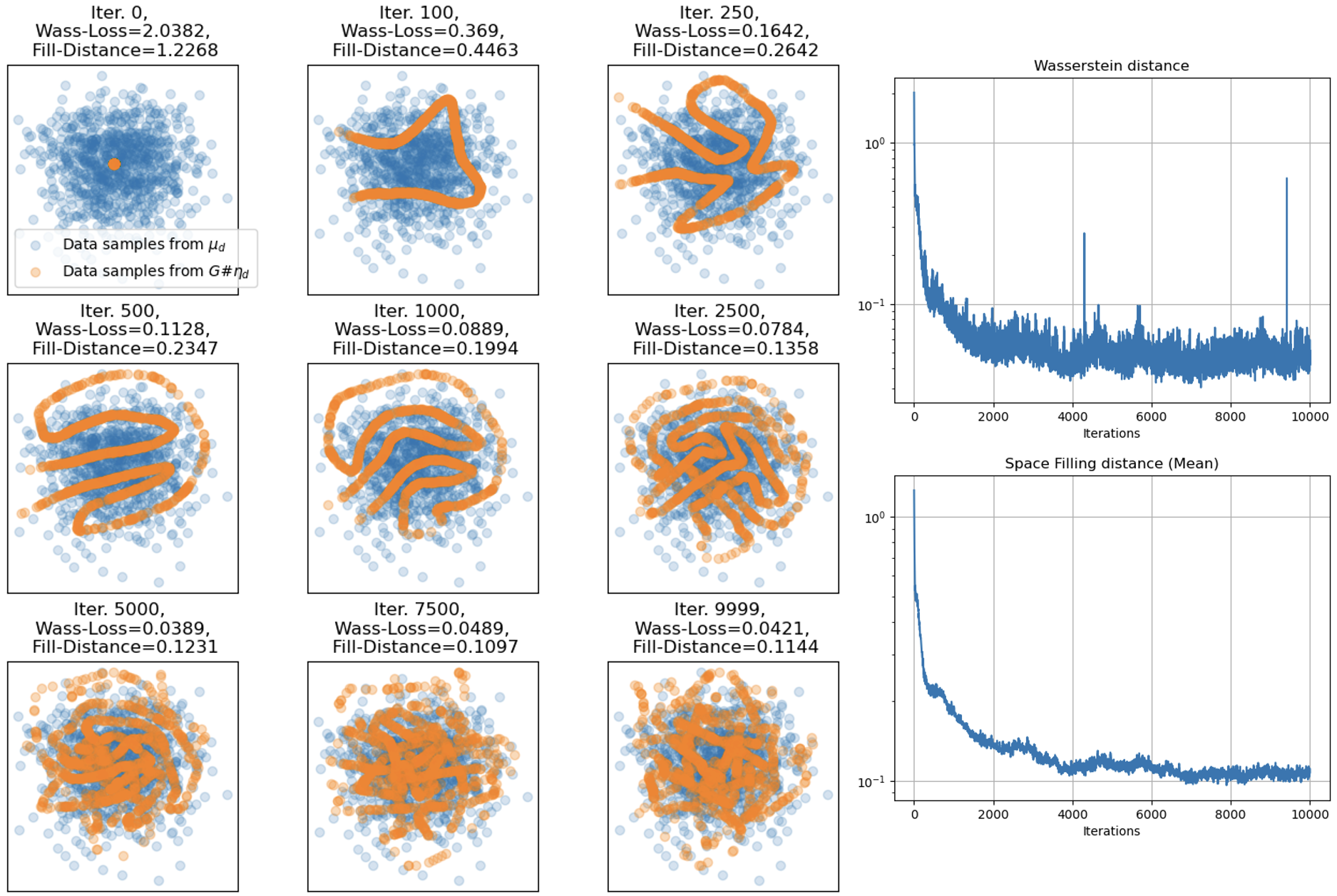}
            \caption{Training results of a 5 hidden layer, 200 node each, fully connected network for 10,000 iterations. Training trajectory (left), Wasserstein loss (top right), and fill distance (bottom right) of mapping a 1-D normally distributed input to a 2-D normally distributed target sample.}
            \label{fig:apdx:sim-1-trajectory-normal-normal}
            % \vspace{-10pt}
        \end{figure}
%\newpage
%\bibliographystyle{chicago}
%\bibliography{ref}

\bibliographystyle{agsm}
\clearpage
\bibliography{ref}

\end{document}